\documentclass{article}

\usepackage[main, preprint]{style}

\usepackage[utf8]{inputenc} 
\usepackage[T1]{fontenc}    
\usepackage[hidelinks]{hyperref} 
\usepackage{url}            
\usepackage{booktabs}       
\usepackage{amsfonts}       
\usepackage{amsmath}        
\usepackage{amssymb}        
\usepackage{nicefrac}       
\usepackage{microtype}      
\usepackage{xcolor}         
\usepackage{graphicx}       
\usepackage{subcaption}     

\definecolor{linkblue}{HTML}{1A56DB}
\newcommand\blfootnote[1]{%
  \begingroup
  \renewcommand\thefootnote{}\footnote{#1}%
  \addtocounter{footnote}{-1}%
  \endgroup
}

\title{Reasoning Models Don't Just Think Longer,\\
They Move Differently}

\author{
\textbf{Anders Gj{\o}lbye}$^{1,2}$ \hspace{0.7cm}
\textbf{Lars Kai Hansen}$^{1}$ \hspace{0.7cm}
\textbf{Sanmi Koyejo}$^{2}$ \\
$^1$Technical University of Denmark \hspace{0.7cm}
$^2$Stanford University \\
\texttt{gjoelbye@cs.stanford.edu} \hspace{0.7cm}
\texttt{lkai@dtu.dk} \hspace{0.7cm}
\texttt{sanmi@cs.stanford.edu}
}

\begin{document}

\maketitle
\blfootnote{%
  \hspace*{-1.8em}%
  \begin{tabular}{@{}c@{\hspace{0.5em}}l@{}}
    \raisebox{-0.18em}{\includegraphics[height=0.95em]{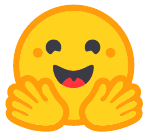}} &
      \href{https://huggingface.co/datasets/gjoelbye/cot-hidden-state-trajectories}{%
        \textcolor{linkblue}{\nolinkurl{https://huggingface.co/datasets/gjoelbye/cot-hidden-state-trajectories}}} \\[2pt]
    \raisebox{-0.18em}{\includegraphics[height=0.95em]{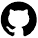}} &
      \href{https://github.com/gjoelbye/reasoning-trajectory-geometry}{%
        \textcolor{linkblue}{\nolinkurl{https://github.com/gjoelbye/reasoning-trajectory-geometry}}} \\
  \end{tabular}%
}

\begin{abstract}
Reasoning-trained language models often spend more tokens on harder problems, but longer chains of thought do not show whether a model is merely computing for more steps or following a different internal trajectory. We study this distinction through hidden-state trajectories during chain-of-thought generation across competitive programming, mathematics, and Boolean satisfiability. Raw trajectory geometry is strongly shaped by generation length: longer generations mechanically alter path statistics, so difficulty-dependent comparisons are misleading without adjustment. After residualizing trajectory statistics on length, difficulty remains systematically coupled to corrected trajectory geometry across all domains studied. The clearest reasoning-specific separation appears in the code domain, where harder problems show more direct corrected trajectories and less heterogeneous local curvature in reasoning-trained models than in matched instruction-tuned baselines. Corrected difficulty-geometry coupling is weaker, but still present, in mathematics and Boolean satisfiability. Prompt-stage linear probes do not mirror the code-domain separation, and behavioral annotations show that stronger corrected coupling co-occurs with strategy shifts and uncertainty monitoring. Together, these findings establish length correction as a prerequisite for generation-time trajectory analysis and show that reasoning training can be associated with distinct corrected trajectory geometry, with the strength of the effect depending on the domain.
\end{abstract}


\section{Introduction}
\label{sec:intro}

Reasoning-trained LLMs often spend more test-time compute on harder problems, producing substantially longer chains of thought and sometimes thousands of unnecessary tokens on easy ones~\citep{chen2025overthinking, wang2025underthinking, snell2024scaling}. Longer traces, however, do not reveal whether a model is merely computing for more steps or following a different internal path. Output length alone cannot distinguish these possibilities: a model may extend the same process for longer, or its hidden-state trajectory may change systematically with problem difficulty.

\begin{figure*}[t]
\centering
\includegraphics[width=1.0\textwidth]{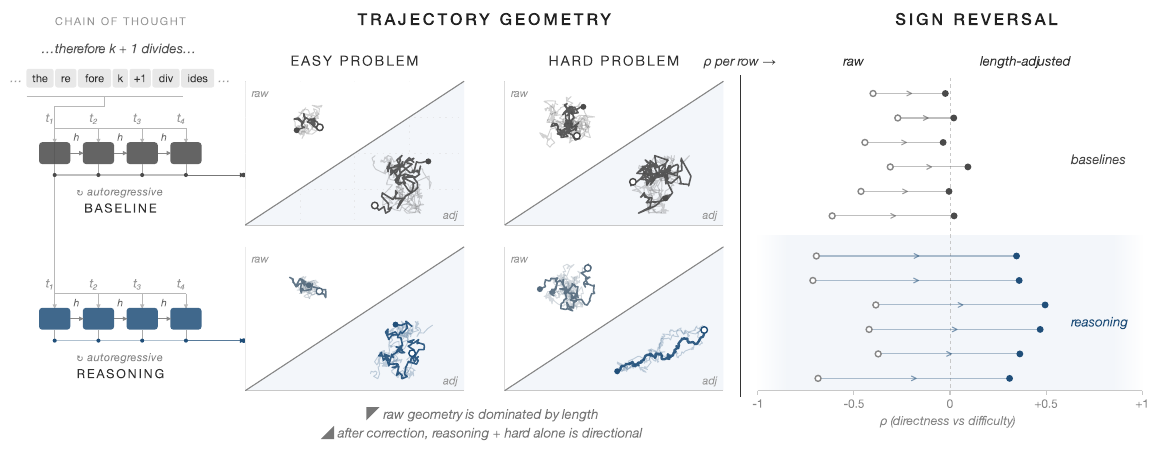}
\caption{Hidden-state trajectory geometry during chain-of-thought generation. Left, autoregressive hidden-state trajectories are extracted from matched reasoning and non-reasoning models on the same problem. Center, raw trajectory geometry is dominated by generation length: longer trajectories, which are more common on harder problems, appear mechanically less direct regardless of model type. Right, Codeforces illustrates the main reasoning-baseline contrast: raw directness-difficulty correlations are negative across models, while length-adjusted correlations separate reasoning models from matched baselines. Full cross-domain results for code, math, and SAT are reported in Figure~\ref{fig:sign_reversal}.}
\label{fig:overview}
\end{figure*}

This distinction matters for interpreting reasoning training. If reasoning models differ from their baselines mainly by allocating more test-time compute, then recent gains may largely reflect better control over computation amount. If difficulty remains coupled to trajectory shape after accounting for length, then reasoning training may also be associated with changes in how computation unfolds during generation. Existing work has mostly studied this issue through outputs, test-time compute allocation, and failure modes such as overthinking or underthinking~\citep{chen2025overthinking, wang2025underthinking, snell2024scaling}. We instead ask whether difficulty-dependent differences are visible in hidden-state trajectories during chain-of-thought generation.

The central complication is that the most intuitive geometric signal is also the easiest to misread. Longer paths are mechanically less direct, a phenomenon well characterized in movement ecology~\citep{benhamou2004tortuosity, codling2008random} but largely unaddressed in generation-time analyses of LLM representations. Since harder problems also elicit longer generations, raw geometry can make hard problems appear less organized simply because their trajectories contain more steps. To avoid this confound, we calibrate item difficulty with an IRT model, extract hidden-state trajectories over generated solution segments, and measure within-model difficulty-geometry coupling after residualizing trajectory statistics on generation length. We then compare this corrected coupling across matched reasoning and instruction-tuned baseline models.

This length-corrected view changes the qualitative result. Before correction, harder problems tend to have less direct trajectories. After residualizing on generation length, the relationship reverses across competitive programming, mathematics, and Boolean satisfiability: harder problems elicit more direct corrected trajectories. The reversal is therefore not only a code-domain effect, but a cross-domain warning that raw generation-time geometry must be interpreted with explicit length correction.

Corrected geometry also separates model classes, but unevenly across domains. In competitive programming, all six reasoning-trained models show positive corrected directness-difficulty coupling, while matched baselines remain near zero (reasoning median $\rho_\perp^D = +0.41$, baseline $-0.04$). In mathematics, the separation is weaker and more heterogeneous ($+0.05$ vs.\ $-0.11$). In Boolean satisfiability, both reasoning and baseline models show positive corrected coupling (medians $+0.26$ and $+0.24$), indicating that corrected difficulty-geometry coupling can also emerge in instruction-tuned baselines. The clearest reasoning-specific contrast is therefore in competitive programming.

Two additional analyses help interpret this geometric signal. First, prompt-stage linear probes do not show the same reasoning-baseline separation as corrected geometry in the code domain, suggesting that the effect is not simply stronger linear access to difficulty before generation. Second, sentence-level behavioral annotations from independent LLM judges show that stronger geometric coupling co-occurs with strategy shifting and uncertainty monitoring. These behavioral analyses are descriptive rather than causal, since the annotations and geometry are measured from the same generated traces.

Our contributions are: \textbf{(i)} identifying generation length as a structural confound in generation-time trajectory geometry; \textbf{(ii)} introducing a length-corrected analysis showing that difficulty remains coupled to corrected geometry across competitive programming, mathematics, and Boolean satisfiability; \textbf{(iii)} showing that reasoning-specific separation is domain-dependent, clearest in competitive programming, while corrected difficulty--geometry coupling persists more weakly elsewhere; \textbf{(iv)} relating the signal to probes and observable reasoning behaviors: linear difficulty decodability does not track the code-domain separation, while stronger corrected coupling co-occurs with strategy shifting and uncertainty monitoring; and \textbf{(v)} a large-scale public trajectory archive pairing generated chain-of-thought traces with sampled generation-time hidden-state trajectories for the matched reasoning and instruction-tuned models.
\section{Related Work}
\label{sec:related}

This paper lies at the intersection of three lines of work: geometric analyses of internal representations, studies of difficulty in LLMs, and work on difficulty-dependent reasoning behavior at inference time.

\paragraph{Trajectory geometry in LLMs.}
Recent work has used trajectory geometry to study the structure of computation in LLM representations. \citet{hosseini2023straighten} showed that LLMs progressively straighten sentence-level trajectories across layers, paralleling temporal straightening in biological neural systems. \citet{zhou2026geometry_reasoning} formalized reasoning as geometric flows in representation space, showing that curvature captures logical structure under carrier-invariant designs. \citet{damirchi2026tat} found that full displacement vectors across layers outperform scalar kinematic descriptors for predicting reasoning validity. These studies establish geometry as a useful lens on internal computation, but they focus on \emph{fixed-depth} trajectories across layers for a single token. Our setting is different: we study generation-time trajectories across tokens at a fixed layer, where path length varies systematically across examples. This makes generation length a central methodological concern, since geometric metrics can change mechanically with trajectory length.  \citet{sun2026reasoning_trajectories} characterize reasoning as trajectories through step-specific representation subspaces, showing that correct and incorrect solutions diverge at late steps and that trajectory-based steering can redirect reasoning.  Our question is complementary: we study token-time trajectories at a fixed layer rather than layer-indexed step representations, and ask whether problem difficulty modulates trajectory geometry after removing the mechanical effects of generation length, a confound not addressed in step-indexed analyses.

\paragraph{Difficulty in LLMs.}
A separate line of work studies how LLMs encode or measure problem difficulty. Linear probes can decode difficulty from hidden states with high accuracy~\citep{lugoloobi2025encode}. IRT has also been adopted for LLM benchmarking and evaluation~\citep{polo2024tinybenchmarks, zhou2025lost_benchmarks, xu2025lart}. \citet{zhu2025llm_already_knows} estimated model-perceived difficulty from hidden representations via a value-function framework, while \citet{lee2025probing_difficulty} identified attention heads with distinct activation patterns for easy versus hard problems. These works show that difficulty is represented in model internals and can be measured continuously. Our goal, however, is not to show that difficulty is encoded, but to use a continuous difficulty variable to study how internal computation changes across problems.

\paragraph{Difficulty-dependent reasoning behavior.}
Work on overthinking, underthinking, and inference-time compute has shown that reasoning models allocate computation differently across easy and hard problems. \citet{snell2024scaling} showed that optimal compute allocation depends on difficulty. \citet{chen2025overthinking} documented overthinking on easy problems, while \citet{wang2025underthinking} identified underthinking on hard problems; \citet{su2025between} showed that both behaviors can coexist. \citet{huang2025manifold_steering} linked overthinking to a low-dimensional activation manifold and proposed steering-based mitigation. These works primarily characterize difficulty-dependent adaptation through outputs or pathological regimes. Our paper asks the complementary internal question: whether reasoning training changes the geometry of the generation-time trajectory itself, across the full difficulty continuum and after controlling for response length.

Taken together, these literatures motivate geometry, difficulty, and inference-time adaptation as relevant lenses, but leave open whether reasoning training changes generation-time internal dynamics as a function of problem difficulty once the strong response-length confound is removed.

\section{Experimental Setup}
\label{sec:setup}

We use a matched design to separate four quantities that are otherwise entangled: problem difficulty, generation length, model class, and trajectory geometry. We define comparable item sets across three domains, calibrate a continuous difficulty scale within each domain, compare matched reasoning and instruction-tuned model pairs on the same items, and extract hidden-state trajectories from generated solution segments.

\textbf{Datasets.}
We evaluate on 500 Easy2Hard-Bench competitive-programming problems~\citep{ding2024easy2hard}, 500 MATH problems~\citep{hendrycks2021math}, and 500 SATBench problems~\citep{satbench}. SATBench items are stratified into five clause-count bins spanning 4--45 clauses and are approximately balanced between satisfiable and unsatisfiable instances within each bin. This yields 1{,}500 items across competitive programming, mathematics, and Boolean satisfiability.

\textbf{Difficulty calibration.}
Native difficulty labels are platform-specific (Codeforces Glicko-2 ratings), coarsely ordinal (MATH levels 1--5), or structural (SAT clause counts; SATBench clause count is the dominant proxy for instance hardness in the synthetic regime studied here). To obtain a continuous latent difficulty scale within each domain, we fit a Rasch model~\citep{rasch1960probabilistic} with a binomial likelihood over repeated runs:
\begin{equation}
\label{eq:irt}
k_{ij} \sim \mathrm{Binomial}\bigl(n_{ij},\; \sigma(\theta_j - b_i)\bigr),
\end{equation}
where $k_{ij}$ is the number of correct completions by model $j$ on item $i$, and $b_i$ is item difficulty. IRT is calibrated separately per domain from 32 models and validated against external labels: Spearman $\rho=0.55$ with Codeforces ratings, $\rho=0.42$ with MATH levels, and $\rho=0.56$ ($r=0.59$) with SAT clause counts. We use $b_i$ as the continuous independent variable throughout. Appendix~\ref{app:irt_details} reports calibration diagnostics, external-label agreement, 1PL--2PL comparisons, and leave-one-out recalibration checks.

\textbf{Matched model pairs.}
The core analysis uses six matched reasoning-baseline comparisons across Qwen, Llama, and Phi families, with three reasoning-training recipes: R1 distillation, SFT+RL, and o3-mini distillation. These six comparisons contain five unique baseline models because Qwen2.5-32B-Instruct serves as the shared baseline for both R1-Distill-Qwen-32B and QwQ-32B. Pair-level counts use six matched comparisons; unique-baseline counts use five baseline models. We state which convention is used wherever counts are reported. The 32B shared-base comparison is especially clean because R1-Distill-Qwen-32B and QwQ-32B differ in reasoning-training recipe while sharing the same instruction-tuned baseline.

\begin{table}[t]
\centering
\caption{Matched model pairs used in the main comparison.}
\label{tab:model_pairs}
\small
\begin{tabular}{@{}llll@{}}
\toprule
Reasoning Model & Baseline & Family & Training \\
\midrule
R1-Distill-Qwen-7B   & Qwen2.5-7B-Instruct   & Qwen  & SFT distillation (R1) \\
R1-Distill-Qwen-14B  & Qwen2.5-14B-Instruct  & Qwen  & SFT distillation (R1) \\
R1-Distill-Qwen-32B  & Qwen2.5-32B-Instruct  & Qwen  & SFT distillation (R1) \\
R1-Distill-Llama-8B  & Llama-3.1-8B-Instruct & Llama & SFT distillation (R1) \\
QwQ-32B              & Qwen2.5-32B-Instruct  & Qwen  & SFT + RL \\
Phi-4-Reasoning      & Phi-4                 & Phi   & SFT distillation (o3-mini) \\
\bottomrule
\end{tabular}
\vspace{-1em}
\end{table}

\textbf{Trajectory extraction overview.}
We extract hidden states at five evenly spaced layers for five runs per problem per model, with 30 runs for R1-Distill-Qwen-7B in stability analyses. Unless otherwise stated, main figures report the median statistic across these five prespecified sampled layers; layer-specific results are reported in Appendix~\ref{app:layer_sensitivity}. We distinguish three representational levels: prompt-stage representations measured at the final prompt token before generation, generation-time trajectories measured over the generated solution segment, and output-level behavior measured from generated traces and correctness outcomes. Correctness is evaluated by code execution for competitive programming, symbolic matching of boxed answers for MATH, and pattern matching of \texttt{SATISFIABLE}/\texttt{UNSATISFIABLE} markers against the ground-truth label for SAT.

\textbf{Trajectory archive.}
The sampled-trajectory archive and the analysis code that reproduces every table and figure are publicly available.\footnote{See Appendix~\ref{app:data_code_availability}.}
The approximately 3\,TB archive pairs generated chain-of-thought traces with sampled generation-time hidden-state trajectories for the matched reasoning and instruction-tuned models, indexed by item, model, run, layer, and token position.
To our knowledge, this is the first large-scale public resource pairing generated reasoning traces with generation-time hidden-state trajectories across matched reasoning and instruction-tuned models.
\section{Trajectory Geometry and Length Correction}
\label{sec:methodology}

\textbf{Generated solution segments.}
For problem $i$, model $m$, run $r$, layer $\ell$, and generation step $t$, let $\mathbf{h}_{imr,t}^{(\ell)}\in\mathbb{R}^d$ be the post-block residual-stream output at the final generated-token position. We restrict analysis to the generated solution segment. For reasoning-trained models with explicit thinking delimiters, this segment is the delimited thinking block. For instruction-tuned baselines, no native thinking delimiter exists, so we define the generated solution segment as the pre-answer text before the first detected answer boundary. In code, the boundary is the first code fence; in math, the first final-answer marker such as \verb|\boxed{}|; in SAT, the first \texttt{SATISFIABLE}/\texttt{UNSATISFIABLE} marker; XML answer tags are fallback markers. If no boundary is detected, the full baseline output is treated as the pre-answer segment. For tagged reasoning models, malformed or missing thinking delimiters are treated as answer-only and therefore produce zero-length reasoning segments; empirically, tagged-but-empty cases were not observed in any reasoning model/domain setting. Because reasoning models often provide explicit thinking delimiters whereas baselines require inferred answer boundaries, Appendix~\ref{app:boundary_segmentation} reports boundary-detection rates, fallback rates, and boundary-policy sensitivity checks.

We use $N_{imr}$ for raw generated solution-segment token length and $T_{imr}$ for sampled trajectory length after stride-based hidden-state sampling. All main trajectory analyses use stride 10. Runs with too few sampled states for a statistic are excluded from that statistic, with exclusion rates reported alongside the segmentation diagnostics. Curvature-based statistics require at least three sampled states.

\textbf{Directness.}
For trajectory $(\mathbf{h}_{imr,0}^{(\ell)},\dots,\mathbf{h}_{imr,T_{imr}}^{(\ell)})$, define path length and net displacement:
\[
L_{imr}^{(\ell)}:=\sum_{t=1}^{T_{imr}}\left\|\mathbf{h}_{imr,t}^{(\ell)}-\mathbf{h}_{imr,t-1}^{(\ell)}\right\|_2,
\quad
\Delta_{imr}^{(\ell)}:=\left\|\mathbf{h}_{imr,T_{imr}}^{(\ell)}-\mathbf{h}_{imr,0}^{(\ell)}\right\|_2.
\]
Directness is
\[
D_{imr}^{(\ell)}=\Delta_{imr}^{(\ell)}/L_{imr}^{(\ell)}\in[0,1].
\]
Directness is our primary interpretable statistic: it measures endpoint efficiency relative to the path actually taken.

\textbf{Curvature variability.}
Let $\kappa_{imr,t}^{(\ell)}$ be Menger curvature over consecutive triples. We define curvature variability as
\[
V_{imr}^{(\ell)}:=\operatorname{sd}(\kappa_{imr,1}^{(\ell)},\dots,\kappa_{imr,T_{imr}-1}^{(\ell)}).
\]
Curvature variability is a robustness-oriented local descriptor. Whereas directness summarizes endpoint efficiency, curvature variability measures heterogeneity in local bending and is less tied to a single net displacement. Negative $\rho_\perp^V$ indicates less heterogeneous local bending after length correction, not necessarily less total turning. As auxiliary checks, we also analyze two intrinsic-dimensionality metrics of the same trajectories, TwoNN and PCA90, using the same raw-versus-corrected correlation framework. Full setup and interpretation are reported in Appendix~\ref{app:auxiliary_geometry}.

\textbf{Length-residualized difficulty-geometry coupling.}
\label{sec:confound}
Within each domain, model, and sampled layer, we average over runs to obtain $\bar D_{im}^{(\ell)}$ and $\bar V_{im}^{(\ell)}$ for each item. Let $b_i$ be the domain-specific IRT difficulty and $N_{im}$ the mean solution-segment token length. We fit a length-only regression separately for each model-layer pair:
\begin{equation}
\label{eq:length_regression}
\bar D_{im}^{(\ell)}=\beta_{0m}^{(\ell)}+\beta_{1m}^{(\ell)}\log N_{im}+\varepsilon_{im}^{(\ell)}.
\end{equation}
Define the residualized component $D_{\perp,im}^{(\ell)}:=\bar D_{im}^{(\ell)}-\hat D_{\parallel,im}^{(\ell)}$, where $\hat D_{\parallel,im}^{(\ell)} = \hat\beta_{0m}^{(\ell)} + \hat\beta_{1m}^{(\ell)} \log N_{im}$ is the OLS-fit length component from Eq.~\ref{eq:length_regression}. Our primary estimand is
\begin{equation}
\label{eq:rho_perp}
\rho_{\perp,m}^{D,(\ell)}:=\rho_S\!\left(b_i, D_{\perp,im}^{(\ell)}\right).
\end{equation}
The resulting statistic $\rho_\perp^D$ measures difficulty-geometry coupling during generation after removing the fitted length component. We apply the same residualization procedure to curvature variability to obtain $\rho_\perp^V$.

Interpretation depends on separating geometry from length, so Appendix~\ref{app:alternative_length_corrections} reports alternative residualizations and length-matched analyses, including $D\sim N^{-1/2}$, $\log D\sim\log N$, length-binned matching, and $T$-based variants using sampled trajectory length rather than raw token length. These checks address the functional form of the length correction; segmentation and layer sensitivity are reported separately in Appendices~\ref{app:boundary_segmentation} and~\ref{app:layer_sensitivity}.

\textbf{Prompt-stage difficulty decodability.}
To assess whether generation-stage coupling is mirrored by stronger linear difficulty information before generation, we extract the hidden state at the final prompt token and train Ridge probes to predict IRT difficulty. For each domain and model-layer pair, probes are trained and evaluated on held-out item splits using the same targets $b_i$; the resulting cross-validated prediction score is used as the prompt-stage decodability measure. We average over runs to obtain one row per item before training, then use 5-fold cross-validation by item. Prompt-stage decodability and generation-stage geometric coupling are different estimands: prompt probes measure linear accessibility of difficulty information before generation, while $\rho_\perp^D$ measures how difficulty is coupled to trajectory shape during generation. Their comparison is diagnostic rather than causal.
\begin{figure*}[t]
\centering
\includegraphics[width=\textwidth]{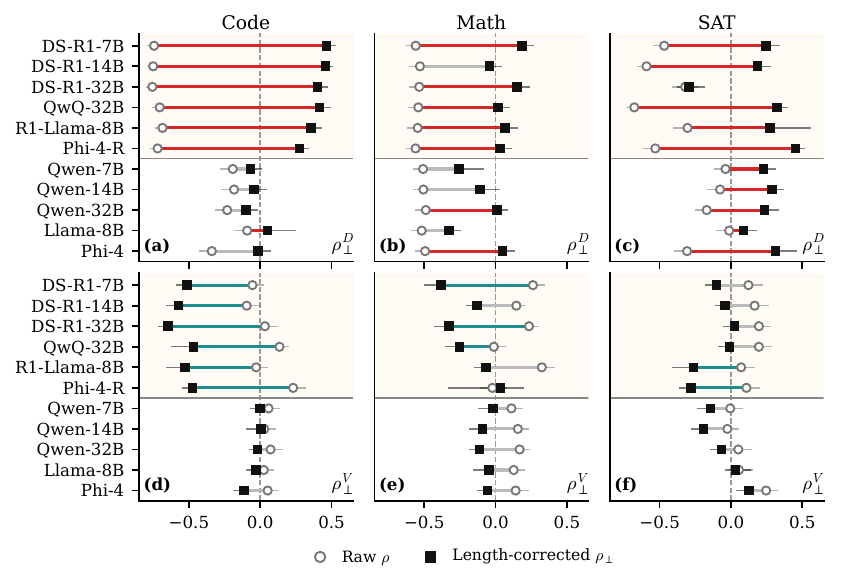}
\vspace{-2em}
\caption{Length correction reveals a sign reversal across all three domains. Hollow circles show raw Spearman correlations with IRT difficulty; filled squares show length-corrected correlations $\rho_\perp$ after residualizing on $\log N$. Panels (a--c) show directness ($\rho_\perp^D$), and panels (d--f) show curvature variability ($\rho_\perp^V$). The sign reversal is cross-domain; the reasoning-baseline separation is strongest on Codeforces and attenuated on SAT.}
\label{fig:sign_reversal}
\vspace{-2em}
\end{figure*}

\section{Results}
\label{sec:results}

Generation length qualitatively changes the interpretation of trajectory geometry. In raw trajectories, harder problems appear less direct because they elicit longer generations, and longer sampled paths are mechanically less direct. After residualizing trajectory statistics on length, this relationship reverses across Codeforces, MATH, and SAT: harder problems have more direct corrected trajectories. The model-class effect is more specific. Corrected geometry separates reasoning models from matched baselines most clearly on Codeforces, weakly on MATH, and only modestly on SAT, where instruction-tuned baselines also show positive corrected coupling.

\subsection{Length Correction Reveals a Cross-Domain Sign Reversal}
\label{sec:sign_reversal}

Figure~\ref{fig:sign_reversal} shows the raw and length-corrected directness-difficulty correlations. On Codeforces, all six reasoning models move from strongly negative raw coupling to positive corrected $\rho_\perp^D$ (median $-0.73 \rightarrow +0.41$), while matched baselines remain near zero after correction (median $-0.04$). On MATH, the same pattern is weaker: corrected medians are $+0.05$ for reasoning models and $-0.11$ for baselines. On SAT, the reversal persists but is not reasoning-specific, with positive corrected medians for both reasoning models and baselines ($+0.26$ and $+0.24$). Per-model 95\% bootstrap CIs are reported in Appendix~\ref{app:alternative_length_corrections}. Thus, SAT extends the length-correction result beyond code while bounding the reasoning-specific interpretation.

Codeforces provides the clearest controlled contrast. R1-Distill-Qwen-32B and QwQ-32B share Qwen2.5-32B-Instruct as their baseline but differ in reasoning-training recipe; both reasoning models show positive corrected directness coupling, while the shared baseline remains near zero. This within-base comparison supports the code-domain separation without relying only on family-level differences.

Curvature variability gives a complementary signal. On Codeforces, reasoning models shift to strongly negative corrected $\rho_\perp^V$ (median $-0.52$), while baselines remain near zero, consistent with less heterogeneous local bending on harder code problems after length correction. MATH shows small effects for both groups, and SAT is intermediate (reasoning median $-0.07$, baseline median $-0.07$). Directness is the more interpretable statistic; curvature variability is the more stable robustness-oriented complement. Appendix~\ref{app:auxiliary_geometry} shows that TwoNN and PCA90 are also length-confounded, but their corrected patterns are weaker and less aligned with the reasoning-baseline contrast.

\subsection{Geometry Gaps Are Not Mirrored by Linear Difficulty Probes}
\label{sec:dynamic_construction}

\begin{figure}[t]
\centering
\includegraphics[width=\linewidth]{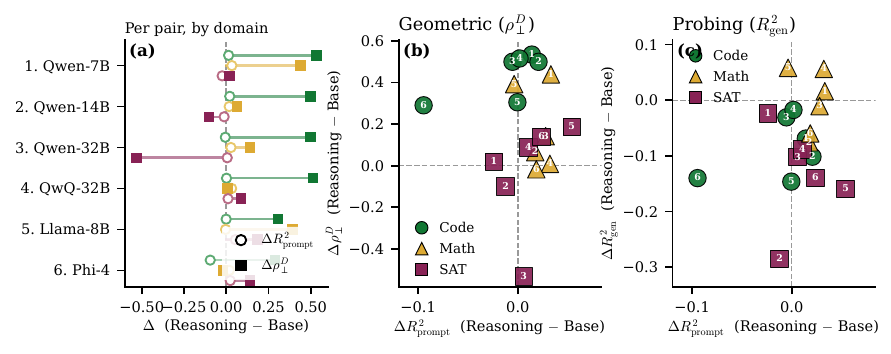}
\vspace{-1em}
\caption{Corrected geometry gaps are not mirrored by stronger linear difficulty decodability. Each point or row is one matched reasoning-baseline pair in one domain, for 18 pair-domain records. All quantities are reasoning minus baseline. Panel (a) compares prompt-stage linear decodability gaps $\Delta R^2_{\mathrm{prompt}}$ with corrected geometric gaps $\Delta\rho_\perp^D$ on a common signed axis. Panel (b) plots $\Delta R^2_{\mathrm{prompt}}$ against $\Delta\rho_\perp^D$; code pairs have large positive geometry gaps but near-zero prompt-probe gaps. Panel (c) plots $\Delta R^2_{\mathrm{prompt}}$ against generation-stage probe gaps $\Delta R^2_{\mathrm{gen}}$ from the peak layer$\times$position heatmap. Linear probing does not show a corresponding reasoning-model advantage, while the corrected geometry gap is largest on Codeforces and smaller on MATH and SAT.}
\label{fig:prompt_generation_dissociation}
\vspace{-1em}
\end{figure}

Figure~\ref{fig:prompt_generation_dissociation} compares corrected geometry gaps with linear difficulty decodability gaps. In code, $\Delta\rho_\perp^D$ is uniformly positive across matched pairs, whereas $\Delta R^2_{\mathrm{prompt}}$ remains near zero and changes sign. The same code pairs also have negative $\Delta R^2_{\mathrm{gen}}$, so the corrected geometric separation is not accompanied by stronger linear difficulty decoding during generation. These results do not rule out nonlinear difficulty representations or differences in how difficulty information is used; they show only that the geometry gap is not a restatement of stronger linear decodability.

Temporal-prefix analyses further show that the code-domain coupling is already visible by the first 10\% of the generated solution segment and is then maintained. In MATH, the signal is weaker and more heterogeneous, and when present it appears to build more gradually. These analyses locate when the corrected geometry signal appears, but do not identify a causal mechanism.

\subsection{Robustness and Scope}
\label{sec:robustness}

The Codeforces reasoning-baseline separation persists across several checks, with metric-dependent caveats. The $\rho_\perp^D$ gap remains under the four length-correction families in Appendix~\ref{app:alternative_length_corrections}, although its magnitude varies; the primary $\log N$ correction agrees closely with the $N^{-1/2}$ correction, while log-log and binned corrections are less stable for directness. Curvature variability is less intuitive but more consistent across correction choices, and therefore provides a useful complement. Residual diagnostics show little remaining length dependence for directness, TwoNN, and PCA90 under the primary correction, but moderate residual length dependence for curvature variability.

Additional checks support the main code-domain sign pattern. Boundary-policy variants indicate that the result is not driven by answer-boundary heuristics (Appendix~\ref{app:boundary_segmentation}); conditioning on correctness preserves the main reasoning-baseline gap within both correct and incorrect subsets, although correctness composition still covaries with difficulty (Appendix~\ref{app:correctness_conditioning}); and layer analyses show that the code-domain signal is present across the five sampled layers (Appendix~\ref{app:layer_sensitivity}). The remaining limitations are that the analyses identify a robust geometric association rather than a robust causal mechanism. Per-domain length-correction values for MATH and SAT are reported in Appendix~\ref{app:alternative_length_corrections}.
\section{Observable Reasoning Behaviors Co-vary with the Geometric Signal}
\label{sec:behavior}

The probe analyses show that the corrected geometry gap is not simply a proxy for stronger linear difficulty decodability. We therefore ask whether the same signal has a visible counterpart in the generated reasoning traces. We focus on Codeforces, where the reasoning-baseline separation in corrected trajectory geometry is strongest; representative annotated traces are visualized in Appendix~\ref{app:behavior}.

We annotate generated solution segments sentence by sentence using three independent LLM judges. The labels cover strategy shifting, uncertainty monitoring, self-correction, verification, problem restatement, and subgoal decomposition, and are aggregated into per-problem behavior rates over the generated solution segment. Judge identities, prompts, metadata visibility, and agreement statistics are reported in Appendix~\ref{app:judge_agreement}; residualized indirect-effect analyses are reported in Appendix~\ref{app:behavior_details}.

Strategy shifting and uncertainty monitoring are the strongest behavioral correlates. In all four R1-distilled models on Codeforces, both behaviors have positive residualized indirect effects with bootstrap confidence intervals excluding zero. QwQ-32B shows the same direction more weakly. Phi-4-Reasoning shows a different profile, with verification as the strongest correlate and no positive uncertainty-monitoring effect. This heterogeneity argues against a single universal mechanism; instead, the annotations provide trace-level correlates of the code-domain geometric separation.

These results are descriptive rather than causal. Behaviors and geometry are measured from the same generated traces, and trace content can itself affect trajectory shape. The annotations therefore connect corrected trajectory geometry to recognizable reasoning dynamics, but do not explain what drives the geometric signal.

\section{Discussion}
\label{sec:discussion}

Generation length is a structural variable in generation-time trajectory geometry. Straightness-style path statistics depend on path structure and length, and prior language-model work has shown that trajectory geometry can be informative when the trajectory regime is well specified~\citep{benhamou2004tortuosity, hosseini2023straighten}. In token-time generation, response length varies with problem difficulty, correctness, and model class, so raw geometric statistics mix trajectory organization with path-length mechanics. Length correction therefore changes the object of analysis: it separates geometry associated with how generation unfolds from geometry induced by how long generation continues.

This length-aware view changes what difficulty-dependent geometry means across the domains we study, and the reasoning-specific contrast is uneven across them. It is strongest in competitive programming, where matched reasoning models and instruction-tuned baselines differ most clearly after length correction, suggesting that reasoning training changes how trajectories adapt as problems become harder; hard code problems more visibly elicit strategy selection, revision, and verification over extended traces, a regime where test-time compute, difficulty, response length, and correctness interact in nontrivial ways~\citep{snell2024scaling, chen2025overthinking, wang2025underthinking, su2025between}. Mathematics and Boolean satisfiability still show corrected difficulty--geometry coupling, but the separation between model classes is weaker. These corrected statistics describe how hidden-state trajectories vary with difficulty during generation; they are not a direct measure of reasoning quality.

The probe and behavioral analyses refine this interpretation. Prompt-stage linear probes do not mirror the code-domain geometry gap, separating difficulty decodability from difficulty-conditioned trajectory dynamics. This makes generation-time geometry a complementary object of study: it asks not only whether difficulty information is present, but how the trajectory evolves while the model produces a solution. Behavioral annotations provide an observable counterpart to the geometric signal. Stronger corrected coupling co-occurs with strategy shifting and uncertainty monitoring, linking the trajectory statistics to recognizable features of generated reasoning traces. These analyses are descriptive rather than causal, since the annotations and geometry are derived from the same outputs. Probe-based interventions did not identify the linear difficulty direction as a causal handle for the corrected geometry signal; details are in Appendix~\ref{app:difficulty_direction_interventions}.

The broader implication is practical. Generation-time representation geometry should be analyzed conditionally on the sampling and segmentation regime. Comparisons between easy and hard problems, correct and incorrect solutions, or reasoning and non-reasoning models should report raw and length-corrected statistics and check residual dependence on length. Without these controls, apparent differences in trajectory organization can reflect path-length mechanics rather than differences in how generation unfolds.
\section{Conclusion}
\label{sec:conclusion}

Generation-time hidden-state geometry cannot be interpreted independently of response length. Across competitive programming, mathematics, and Boolean satisfiability, raw trajectory statistics conflate difficulty-dependent structure with the mechanical effects of longer generations, while length-corrected statistics reveal systematic coupling between item difficulty and trajectory geometry. Within this corrected view, reasoning-trained models show their clearest separation from matched instruction-tuned baselines in competitive programming, where harder problems induce more direct trajectories and less heterogeneous local curvature; the weaker separation in the other domains shows that the effect is domain-dependent rather than a universal signature of reasoning training. These results make length correction a necessary step for studying representation geometry during generation and suggest that reasoning training can change how internal trajectories adapt to problem difficulty. Establishing causal control over these trajectories remains an important direction for understanding how reasoning behavior is organized during generation.

\section*{Limitations}
The main limitations concern segmentation, correction choice, and causal interpretation. Reasoning models with explicit thinking delimiters provide cleaner solution segments than instruction-tuned baselines, whose boundaries must be inferred from answer markers; baseline comparisons are therefore partly dependent on segmentation policy. Directness is more sensitive to the length-correction family than curvature variability, so it should be read together with robustness checks and complementary metrics. Correctness still covaries with difficulty, even though correctness-conditioned analyses preserve the main code-domain pattern. Behavioral annotations provide observable correlates of the geometric signal, but not a mechanism, since labels and geometry come from the same traces. Linear-probe-based interventions along the difficulty direction do not produce dose-dependent geometric changes (Appendix~\ref{app:difficulty_direction_interventions}); identifying the actual driver of difficulty-conditioned trajectory structure remains open. Fixed-stride sampling may also miss finer temporal structure.

\section*{Acknowledgments}
This work was supported by the Novo Nordisk Foundation grant
NNF22OC0076907, ``Cognitive spaces -- Next generation explainability'',
the Pioneer Centre for AI, DNRF grant number P1, and the Danish Data
Science Academy, which is funded by the Novo Nordisk Foundation
(NNF21SA0069429) and VILLUM FONDEN (40516). Anders Gj{\o}lbye
conducted part of while visiting Stanford University. Sanmi
Koyejo acknowledges support by NSF 2046795 and 2205329, IES
R305C240046, ARPA-H, the MacArthur Foundation, Schmidt Sciences, HAI,
OpenAI, Microsoft, and Google.

\bibliographystyle{plainnat}
\bibliography{references}

\newpage
\appendix
\section{Data and Difficulty Calibration}
\label{app:data_calibration}

\subsection{Datasets}

We evaluate on 500 Easy2Hard-Bench competitive-programming problems, 500 MATH problems, and 500 SATBench problems. SATBench items are stratified into five clause-count bins spanning 4--45 clauses and are approximately balanced between satisfiable and unsatisfiable instances within each bin. No item is shared across domains. Native difficulty labels are used for external validation of the latent difficulty scale, not as the primary independent variable in trajectory analyses.

\subsection{Model inventory}
\label{app:models}

Table~\ref{tab:full_models} lists all 32 models. The 11 unique models forming the six matched pairs (Qwen2.5-32B-Instruct serves as the shared baseline of two pairs) have hidden-state access; activations are extracted at five evenly spaced layers with stride 10. The remaining 21 models contribute correctness data only and stabilize the IRT difficulty scale. Per-domain calibration pools contain 32 models on each domain.

\begin{table}[h]
\centering
\caption{All models used in this study. Hidden dimension $d$ and layer count $L$ are shown for the matched-pair models. $\theta_{\text{code}}$, $\theta_{\text{math}}$, $\theta_{\text{sat}}$ are pooled binomial Rasch ability estimates per domain.}
\label{tab:full_models}
\small
\begin{tabular}{@{}llrrlrrr@{}}
\toprule
Model & Source & $d$ & $L$ & Role & $\theta_{\text{code}}$ & $\theta_{\text{math}}$ & $\theta_{\text{sat}}$ \\
\midrule
\multicolumn{8}{@{}l}{\emph{Matched pairs (11 models)}} \\
R1-Distill-Qwen-7B & DeepSeek & 3{,}584 & 28 & Reasoning & $-0.99$ & $+1.57$ & $+0.24$ \\
R1-Distill-Qwen-14B & DeepSeek & 5{,}120 & 48 & Reasoning & $+0.34$ & $+1.75$ & $+0.74$ \\
R1-Distill-Qwen-32B & DeepSeek & 5{,}120 & 64 & Reasoning & $+0.64$ & $+1.66$ & $+1.16$ \\
R1-Distill-Llama-8B & DeepSeek & 4{,}096 & 32 & Reasoning & $-0.71$ & $+1.22$ & $+0.53$ \\
QwQ-32B & Qwen & 5{,}120 & 64 & Reasoning & $+0.82$ & $+1.87$ & $+1.21$ \\
Phi-4-Reasoning & Microsoft & 5{,}120 & 40 & Reasoning & $+0.34$ & $+2.99$ & $+1.38$ \\
Qwen2.5-7B-Instruct & Qwen & 3{,}584 & 28 & Baseline & $-3.28$ & $+1.59$ & $+0.03$ \\
Qwen2.5-14B-Instruct & Qwen & 5{,}120 & 48 & Baseline & $-2.84$ & $+1.88$ & $+0.25$ \\
Qwen2.5-32B-Instruct & Qwen & 5{,}120 & 64 & Baseline & $-1.89$ & $+2.04$ & $+0.31$ \\
Llama-3.1-8B-Instruct & Meta & 4{,}096 & 32 & Baseline & $-4.61$ & $+0.27$ & $-0.19$ \\
Phi-4 & Microsoft & 5{,}120 & 40 & Baseline & $-2.06$ & $+2.05$ & $+0.33$ \\
\midrule
\multicolumn{8}{@{}l}{\emph{IRT calibration models (21)}} \\
Phi-3.5-Mini-Instruct & Microsoft & -- & -- & Calibration & $-4.59$ & $-0.43$ & $+0.02$ \\
Gemma-2-9B-IT & Google & -- & -- & Calibration & $-4.16$ & $-0.19$ & $-0.41$ \\
Mistral-7B-Instruct & Mistral & -- & -- & Calibration & $-5.85$ & $-3.17$ & $-0.13$ \\
Qwen2.5-Math-7B-Instruct & Qwen & -- & -- & Calibration & $-5.66$ & $+2.09$ & $+0.04$ \\
DeepSeek-7B-Chat & DeepSeek & -- & -- & Calibration & $-7.82$ & $-1.94$ & $-0.98$ \\
OLMo-7B-Instruct & AI2 & -- & -- & Calibration & $-7.64$ & $-0.92$ & $-0.04$ \\
Qwen2-7B-Instruct & Qwen & -- & -- & Calibration & $-5.09$ & $+0.68$ & $+0.08$ \\
Zephyr-7B-Beta & HuggingFace & -- & -- & Calibration & $-6.86$ & $-4.52$ & $-0.14$ \\
Mistral-Small-24B & Mistral & -- & -- & Calibration & $-2.86$ & $+1.84$ & $+0.24$ \\
Claude Haiku 4.5 & Anthropic & -- & -- & Calibration & $+0.70$ & $+2.81$ & $+0.82$ \\
Claude Sonnet 4 & Anthropic & -- & -- & Calibration & $+0.65$ & $+3.12$ & $+0.94$ \\
DeepSeek-V3 & DeepSeek & -- & -- & Calibration & $+1.53$ & $+2.87$ & $+2.67$ \\
Gemini 2.5 Flash Lite & Google & -- & -- & Calibration & $+0.65$ & $+3.13$ & $+1.20$ \\
Gemini 2.5 Flash & Google & -- & -- & Calibration & $+1.52$ & $+3.49$ & $+2.07$ \\
Gemini 2.5 Pro & Google & -- & -- & Calibration & $+2.99$ & $+1.02$ & $+3.64$ \\
Gemma-3-27B & Google & -- & -- & Calibration & $-1.20$ & $+2.78$ & $+0.18$ \\
GPT-4o-Mini & OpenAI & -- & -- & Calibration & $-1.59$ & $+1.69$ & $+0.09$ \\
GPT-4o & OpenAI & -- & -- & Calibration & $-2.59$ & $+1.75$ & $+0.27$ \\
Llama-3.3-70B-Instruct & Meta & -- & -- & Calibration & $-1.42$ & $+2.28$ & $+0.23$ \\
o4-mini & OpenAI & -- & -- & Calibration & $+2.60$ & $+2.38$ & $+2.48$ \\
Qwen2.5-72B-Instruct & Qwen & -- & -- & Calibration & $-2.09$ & $+2.03$ & $+0.36$ \\
\bottomrule
\end{tabular}
\end{table}

\subsection{Matched-pair convention}

The main analysis uses six matched reasoning-baseline comparisons. Because Qwen2.5-32B-Instruct is the shared baseline of both R1-Distill-Qwen-32B and QwQ-32B, the six comparisons contain five unique baseline models. Pair-level summaries count six baseline appearances; unique-model summaries count five. We state which convention is used whenever reporting counts.

\subsection{Correctness evaluation}
\label{app:correctness}

\paragraph{Competitive programming.}
The last Python code block is extracted from each trace and executed against official test cases in sandboxed subprocesses (5\,s timeout). Output comparison uses whitespace normalization, float tolerance ($10^{-6}$), and case-insensitive boolean matching.

\paragraph{Mathematics.}
The last \texttt{\textbackslash boxed\{\}} expression is extracted (handling nested braces). Comparison uses exact string matching, SymPy symbolic equivalence, and numerical tolerance.

\paragraph{Boolean satisfiability.}
The first \texttt{SATISFIABLE} or \texttt{UNSATISFIABLE} token in the trace is extracted (case-insensitive, with optional surrounding markup) and compared against the ground-truth label. Traces with no detected marker are treated as incorrect.

\subsection{Prompt templates and decoding}

All models are sampled with temperature $0.6$, nucleus $p=0.95$, a maximum of 32{,}768 tokens, and a fixed seed per run. R1-distilled models receive a \texttt{<think>} prefix to trigger extended reasoning; Phi-4-Reasoning uses its native reasoning prompt format; instruction-tuned baselines receive identical problem statements without thinking delimiters. The main analysis uses five runs per problem per model, with 30 runs for R1-Distill-Qwen-7B in the run-count stability analysis (Appendix~\ref{app:run_count_stability}).

\subsection{Rasch calibration}
\label{app:irt_details}

For item $i$ and model $j$, observed successes are modeled as
\[
k_{ij} \sim \mathrm{Binomial}\!\left(n_{ij}, \sigma(\theta_j - b_i)\right),
\]
where $\theta_j$ is model ability and $b_i$ is item difficulty. We use the fitted $b_i$ as the continuous difficulty variable in all downstream analyses.

\paragraph{Overview.}
We validate pooled binomial Rasch calibration along four axes: boundary-item structure and item-pool coverage, agreement with withheld native difficulty labels, parsimony relative to 2PL, and leave-one-out (LOO) stability under removal of any single calibration model.

\begin{table}[t]
\centering
\caption{Calibration pool and optimization summary.}
\label{tab:irt_optimization}
\small
\begin{tabular}{@{}lccc@{}}
\toprule
 & Code & Math & SAT \\
\midrule
Items & 500 & 500 & 500 \\
Calibration models & 32 & 32 & 32 \\
Observations ($i\times j$) & 16{,}000 & 16{,}000 & 16{,}000 \\
Learning rate / max epochs / patience & 0.05 / 2000 / 200 & 0.05 / 2000 / 200 & 0.05 / 2000 / 200 \\
Stopped epoch & 3{,}104 & 2{,}353 & 1{,}574 \\
Final loss & 25{,}637.12 & 23{,}267.25 & 27{,}346.88 \\
\bottomrule
\end{tabular}
\end{table}

\paragraph{Optimization behavior.}
MAP estimation uses Adam with PyTorch defaults ($\beta_1=0.9, \beta_2=0.999$); full settings are in Table~\ref{tab:irt_optimization}. Across the three domains, optimization runs for 1{,}574--3{,}104 epochs before the loss plateaus. This slow-tail convergence is expected for Rasch MAP fitting and is benign for downstream inference because subsequent analyses use rank-level properties of $b_i$, which are highly stable under recalibration.

\paragraph{Boundary structure and coverage.}
Boundary items are solved on every run by every model (ceiling) or failed on every run by every model (floor). In code, 457 items are informative and 43 are floor-boundary; in math, 470 are informative and 30 are floor-boundary; in SAT, all 500 items are informative with neither floor nor ceiling boundary mass, reflecting that no SATBench instance is solved by every model or failed by every model in the 32-model calibration pool. All three domains have zero ceiling-boundary items.

\begin{figure}[t]
\centering
\includegraphics[width=\textwidth]{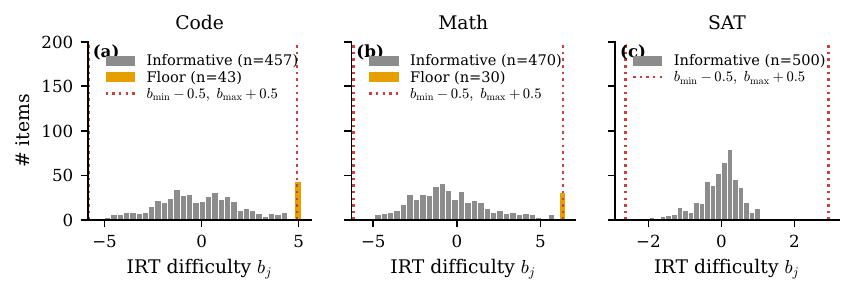}
\caption{Boundary structure of pooled 1PL item difficulties. Left: code domain. Center: math domain. Right: SAT domain. Code and math show floor-only boundary mass (no ceiling items), with boundary difficulties fixed by the prior-offset convention described in Section~\ref{sec:setup}. SAT has no boundary items in the 32-model calibration pool: all 500 SATBench instances are informative.}
\label{fig:boundary_items}
\end{figure}

\paragraph{External validity.}
Against withheld native labels, code-domain difficulty aligns with Codeforces Glicko-2 ratings (Pearson $r=0.518$, Spearman $\rho=0.552$, $n=500$). Math-domain difficulty aligns with MATH levels (Spearman $\rho=0.424$); cross-level differences are strong (Kruskal--Wallis $H=95.56$, 4 d.f., $p<10^{-20}$). SAT-domain difficulty aligns with SAT clause counts (Pearson $r=0.589$, Spearman $\rho=0.559$, $n=500$, $p<10^{-42}$), the strongest external alignment of the three domains.

\begin{figure}[t]
\centering
\includegraphics[width=\textwidth]{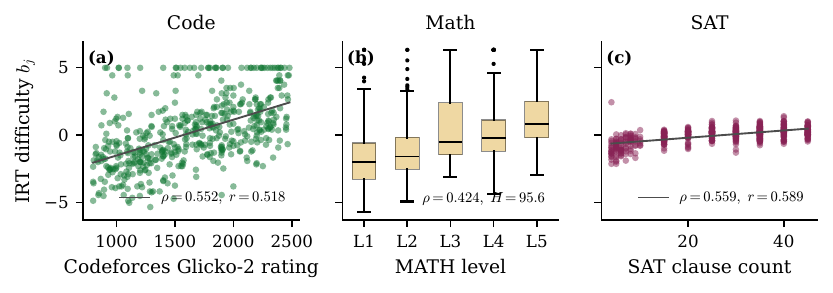}
\caption{External validation using labels not used during IRT fitting. (a) Code: IRT difficulty versus Codeforces Glicko-2 rating. (b) Math: IRT difficulty stratified by MATH levels L1--L5. (c) SAT: IRT difficulty versus SAT clause count (Pearson $r=0.589$, Spearman $\rho=0.559$, $n=500$).}
\label{fig:external_validation}
\end{figure}

\begin{table}[t]
\centering
\caption{Compact validation summary for pooled IRT calibration.}
\label{tab:irt_validation_summary}
\small
\begin{tabular}{@{}llc@{}}
\toprule
Axis & Statistic & Value \\
\midrule
Boundary structure (code) & Informative / floor / ceiling & 457 / 43 / 0 \\
Boundary structure (math) & Informative / floor / ceiling & 470 / 30 / 0 \\
Boundary structure (SAT)  & Informative / floor / ceiling & 500 / 0 / 0 \\
External validity (code) & Pearson $r$, Spearman $\rho$ & 0.518, 0.552 \\
External validity (math) & Spearman $\rho$, Kruskal--Wallis $H$ & 0.424, 95.56 \\
External validity (SAT)  & Pearson $r$, Spearman $\rho$ & 0.589, 0.559 \\
Parsimony (code) & $\rho(1\mathrm{PL},2\mathrm{PL})$ & 0.9892 \\
Parsimony (math) & $\rho(1\mathrm{PL},2\mathrm{PL})$ & 0.9656 \\
Parsimony (SAT)  & $\rho(1\mathrm{PL},2\mathrm{PL})$ & 0.9413 \\
LOO stability (code) & Median / min Spearman $\rho$ & 0.9996 / 0.9906 \\
LOO stability (math) & Median / min Spearman $\rho$ & 0.9995 / 0.9982 \\
LOO stability (SAT)  & Median / min Spearman $\rho$ & 0.9949 / 0.9915 \\
\bottomrule
\end{tabular}
\end{table}

\paragraph{1PL parsimony and LOO stability.}
Allowing free item discriminations in 2PL leaves the difficulty ordering nearly unchanged ($\rho(1\mathrm{PL},2\mathrm{PL})=0.9892$ code, $0.9656$ math, $0.9413$ SAT), supporting 1PL as the parsimonious choice. LOO recalibration over models yields near-identical rankings relative to the full-pool fit (median Spearman $\rho$ of $0.9996$ on code, $0.9995$ on math, and $0.9949$ on SAT; minimum $\rho$ of $0.9906$ code, $0.9982$ math, $0.9915$ SAT), indicating that no single model drives the scale.

External agreement is moderate rather than near-perfect ($\rho\approx 0.42$ to $0.56$), as expected because Codeforces ratings reflect contest-population dynamics, MATH levels are coarse ordinals, and SAT clause counts measure structural rather than algorithmic difficulty. The resulting latent variable is therefore interpreted as model-pooled difficulty, with strong internal stability and meaningful but not identity-level alignment to native labels.

\subsection{Compute and storage.}
Local runs for the larger open-weight models used three NVIDIA H100 GPUs with 80\,GB of VRAM each; smaller open-weight models were run on NVIDIA L40S GPUs. The main cost was not ordinary decoding, but decoding while saving hidden states, which slowed generation substantially. Across the full set of model families, domains, and robustness checks, local generation and activation extraction took several weeks to months of wall-clock time. The stored artifacts occupy approximately 3\,TB (see \S\ref{sec:setup}); the bulk is extracted hidden states and intermediate trajectory representations. API calls were used for non-local calibration models only; hidden-state trajectories were extracted exclusively from locally run open-weight models.

\section{Trajectories and Metrics}
\label{app:trajectory_metrics}

\subsection{Hidden-state tensor}
\label{app:trajectory_extraction}

For each selected decoder layer $\ell$, we extract the post-block residual-stream output at generation step $t$, evaluated at the final generated-token position. During generation, this yields a trajectory
\[
\mathbf{h}_{imr,0}^{(\ell)}, \mathbf{h}_{imr,1}^{(\ell)}, \dots, \mathbf{h}_{imr,T_{imr}}^{(\ell)} \in \mathbb{R}^d.
\]
This corresponds to the direct output of the decoder layer module after the full attention and feed-forward block with residual connections, rather than pre-layernorm, attention-only, or FFN-only activations.

\subsection{Sampling, layers, and stride}

Hidden states are extracted at five evenly spaced layers, with indices $\{\lfloor i \cdot (L{-}1) / 4 \rfloor : i = 0, \ldots, 4\}$, where $L$ is the total number of layers. States are captured every 10 generated tokens. Trajectory-geometry statistics report the median across the five sampled layers, while probe scores (Appendix~\ref{app:probes_interventions}) use the peak (layer $\times$ position) cell selected independently of the trajectory metric. Layer-specific values for $\rho_\perp^D$ are reported in Appendix~\ref{app:layer_sensitivity}.

\subsection{Generated solution segments}

For models with explicit reasoning delimiters, the closing delimiter defines the boundary between reasoning and answer phases. For non-R1 models, the boundary is detected heuristically: the first code fence for competitive programming, \texttt{\textbackslash boxed\{\}} for mathematics, the first \texttt{SATISFIABLE}/\texttt{UNSATISFIABLE} marker for Boolean satisfiability, and XML answer tags as a fallback. If no boundary pattern is detected, the full output is treated as reasoning for baseline models and as answer-only for tagged reasoning models. All trajectory metrics are computed on the generated solution segment.

Segmentation is exact for tagged reasoning models and heuristic for instruction-tuned baselines. This asymmetry is a genuine limitation of current open-model formats. Boundary-detection rates, fallback rates, and boundary-policy sensitivity checks are reported in Appendix~\ref{app:boundary_segmentation}.

\subsection{Trajectory metrics}

\paragraph{Directness.}
For trajectory $(\mathbf{h}_{imr,0}^{(\ell)},\dots,\mathbf{h}_{imr,T_{imr}}^{(\ell)})$, define path length and net displacement
\[
L_{imr}^{(\ell)} := \sum_{t=1}^{T_{imr}} \left\| \mathbf{h}_{imr,t}^{(\ell)} - \mathbf{h}_{imr,t-1}^{(\ell)} \right\|_2,
\quad
\Delta_{imr}^{(\ell)} := \left\| \mathbf{h}_{imr,T_{imr}}^{(\ell)} - \mathbf{h}_{imr,0}^{(\ell)} \right\|_2.
\]
Directness is $D_{imr}^{(\ell)} = \Delta_{imr}^{(\ell)} / L_{imr}^{(\ell)} \in [0,1]$. Runs with fewer than two sampled states cannot define directness and are excluded from directness analyses.

\paragraph{Curvature variability.}
For three consecutive points $A, B, C \in \mathbb{R}^d$ along a trajectory, the Menger curvature is
\begin{equation}
\label{eq:curvature}
\kappa(A, B, C) = \frac{4 \cdot \mathrm{Area}(\triangle ABC)}{|AB| \cdot |BC| \cdot |AC|},
\end{equation}
where the triangle area in $\mathbb{R}^d$ is
\[
\mathrm{Area}(\triangle ABC) = \tfrac{1}{2}\sqrt{\|\mathbf{u}\|^2 \|\mathbf{v}\|^2 - (\mathbf{u}^\top \mathbf{v})^2}, \quad \mathbf{u} = B - A,\; \mathbf{v} = C - A.
\]
For trajectory $\bigl(\mathbf{h}_{imr,t}^{(\ell)}\bigr)_{t=0}^{T_{imr}}$, curvature variability is
\[
V_{imr}^{(\ell)} := \operatorname{sd}\!\left(\kappa_{imr,1}^{(\ell)}, \dots, \kappa_{imr,T_{imr}-1}^{(\ell)}\right).
\]
Curvature variability is defined only for trajectories with at least three sampled states. It measures heterogeneity in local bending rather than total turning. Negative $\rho_\perp^V$ indicates less heterogeneous local bending after length adjustment, not necessarily less total turning.

\paragraph{TwoNN intrinsic dimension.}
We estimate intrinsic dimension from the ratio of the two nearest-neighbor distances of each sampled state~\citep{facco2017twonn}. Let $r_{i,1}$ and $r_{i,2}$ be the distances from state $i$ to its first and second nearest neighbors among the trajectory's sampled states. Then
\[
\mu_i = \frac{r_{i,2}}{r_{i,1}}, \qquad
\widehat d_{\mathrm{TwoNN}} =
\left(\frac{1}{n}\sum_{i=1}^{n}\log \mu_i\right)^{-1},
\]
where $n$ is the number of sampled states with a nonzero first-nearest-neighbor distance. Sampled states with duplicate nearest neighbors ($r_{i,1}=0$) are excluded, and nearest neighbors are computed within the sampled states of the same trajectory.

\paragraph{PCA90 dimensionality.}
PCA90 reports the smallest number of principal components capturing 90\% of the variance of the trajectory's sampled states~\citep{jolliffecadima2016pca}. Let $Z \in \mathbb{R}^{T \times d}$ be the matrix of sampled trajectory states after centering across the $T$ sampled time points, and let $\lambda_1 \geq \cdots \geq \lambda_r$ be the nonzero eigenvalues of the sample covariance, with $r \leq \min(T-1, d)$. Then
\[
\mathrm{PCA90}(Z) =
\min\!\left\{ k : \frac{\sum_{j=1}^{k}\lambda_j}{\sum_{j=1}^{r}\lambda_j} \geq 0.90 \right\}.
\]
Length-residualized difficulty couplings for TwoNN and PCA90 are reported in Appendix~\ref{app:auxiliary_geometry}.

\section{Length Dependence and Auxiliary Geometry}
\label{app:length_geometry}

\subsection{Alternative length corrections}
\label{app:alternative_length_corrections}

The primary correction residualizes each trajectory statistic on $\log N$ separately within domain, model, and sampled layer; the residuals are then correlated with IRT difficulty $b_i$ at the item level. We compare this estimator against $N^{-1/2}$ residualization, log-log residualization, and length-binned matching to assess how strongly the reasoning-baseline contrast depends on the functional form of the length correction.

Under the primary $\log N$ correction, Codeforces $\rho_\perp^D$ for reasoning models is positive for all six matched pairs (median $+0.41$), while matched baselines center near zero or negative (median $-0.04$). The complementary $\rho_\perp^V$ separation is similarly clean (reasoning median $-0.52$, baseline near zero at $-0.02$). On MATH, $\rho_\perp^D$ is mostly near zero after residualization, while $\rho_\perp^V$ shows small negative shifts for most reasoning models. The code-domain reasoning signal is therefore not a raw-length artifact, although its estimated strength depends on correction family for $\rho_\perp^D$.

\begin{table}[t]
\centering
\caption{Cross-method consistency: Spearman agreement with the primary $\log N$ correction.}
\label{tab:length_cross_method}
\small
\begin{tabular}{lcc}
\toprule
Comparison & $\rho_\perp^D$ & $\rho_\perp^V$ \\
\midrule
$\log N$ vs $N^{-1/2}$ & $+0.96$ & $+0.96$ \\
$\log N$ vs log-log    & $-0.003$ & $+0.95$ \\
$\log N$ vs binned     & $-0.62$ & $+0.50$ \\
\bottomrule
\end{tabular}
\end{table}

\begin{figure*}[t]
\centering
\includegraphics[width=\textwidth]{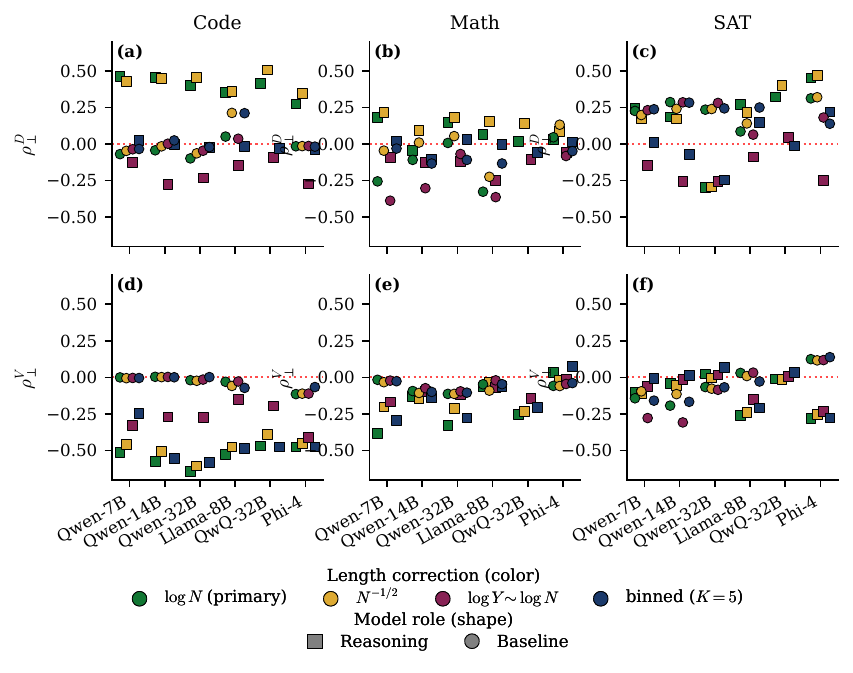}
\caption{Length-correction robustness across four correction families for $\rho_\perp^D$ and $\rho_\perp^V$. Each point shows one model under one correction family. The primary $\log N$ and $N^{-1/2}$ corrections agree closely, while directness is less stable under log-log and binned corrections. Curvature variability is more stable across correction families.}
\label{fig:length_correction}
\end{figure*}

\begin{figure*}[t]
\centering
\includegraphics[width=\textwidth]{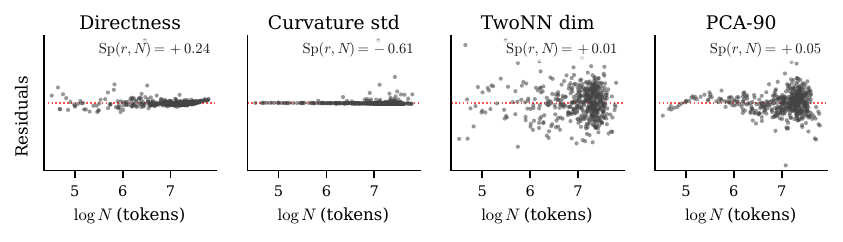}
\caption{Codeforces residual length diagnostics under the primary $\log N$ correction. Each panel plots item-level residuals from the primary length-only regression against $\log N$ for one trajectory statistic. The reported Spearman correlations measure remaining monotonic association with generation length after residualization. Directness, TwoNN, and PCA90 show little residual length association, while curvature variability retains a moderate negative association, indicating that curvature should be interpreted as correction-family-stable rather than fully length-independent.}
\label{fig:residual_length_diagnostics}
\end{figure*}

The $\log N$ and $N^{-1/2}$ corrections agree closely for both $\rho_\perp^D$ and $\rho_\perp^V$ (Spearman $+0.96$ each); agreement is much weaker for directness under log-log and binned corrections ($-0.003$ and $-0.62$), while curvature variability remains more stable across correction families (Codeforces reasoning medians $-0.50, -0.46, -0.25, -0.02$ across the four methods), although Codeforces residual diagnostics under the primary $\log N$ correction (Figure~\ref{fig:residual_length_diagnostics}) show that it retains moderate residual length association. We use the $\log N$ correction as the primary estimator because it is a simple monotone adjustment for generation length, applies uniformly across all trajectory statistics, and agrees closely with the $N^{-1/2}$ correction. Under this primary correction, Codeforces gives the clearest reasoning-baseline separation.

\subsection{Auxiliary dimensionality descriptors}
\label{app:auxiliary_geometry}

As auxiliary descriptors, we apply the same raw-versus-corrected correlation analysis from Figure~\ref{fig:sign_reversal} to TwoNN intrinsic dimension and PCA90 dimensionality. For each model and domain, we compute Spearman correlations with pooled-IRT difficulty before and after residualizing each metric on $\log N$, with 95\% percentile bootstrap confidence intervals (1000 resamples).

The resulting pattern is asymmetric across metrics. TwoNN on Codeforces partially reproduces the directness sign-reversal structure: reasoning rows move from negative raw correlations to near-zero or weakly positive corrected values (median $-0.41 \rightarrow +0.03$), while baselines change little (median $-0.01 \rightarrow +0.07$). PCA90 shows a strong positive raw association with difficulty, consistent with a positive length confound; after correction, code-domain reasoning rows remain only modestly positive (median $+0.65 \rightarrow +0.18$), while baselines center near zero or slightly negative ($+0.24 \rightarrow -0.03$). Math-domain values are mixed across both groups. On SAT, TwoNN raw correlations are near zero for both groups (reasoning median $+0.05$, baseline $+0.09$) and corrected values rise modestly ($+0.11$ R, $+0.17$ B); PCA90 reverses sign relative to code (raw R $+0.58$, B $+0.07$; corrected $0.00$ R, $-0.26$ B).

\begin{figure*}[t]
\centering
\includegraphics[width=\textwidth]{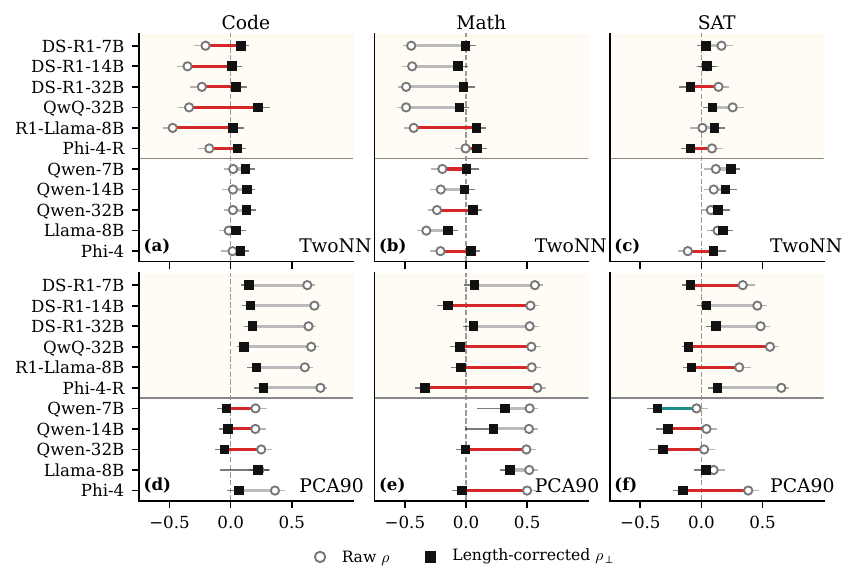}
\caption{Length correction applied to two intrinsic-dimensionality metrics of hidden-state trajectories. Same dumbbell idiom as Figure~\ref{fig:sign_reversal}: each row is one model, hollow circles show raw Spearman $\rho$ with pooled-IRT difficulty, and filled squares show length-corrected $\rho_\perp$ after residualization on $\log N$. Whiskers are 95\% bootstrap confidence intervals (1000 resamples). Panels (a--c) plot TwoNN dimensionality on Codeforces, MATH, and SAT; panels (d--f) plot PCA90 dimensionality on the same three domains. On code, TwoNN shows an attenuated reasoning-model sign-reversal pattern, while PCA90 remains positive but strongly shrunk after correction (reasoning median $+0.18$ vs baseline $-0.03$). Math panels are weaker and mixed. On SAT, TwoNN does not reproduce the directness reversal and PCA90 reverses sign relative to code.}
\label{fig:dimensionality_signal}
\end{figure*}

These dimensionality descriptors show that length affects more than the directness ratio, but their corrected patterns are weaker and less aligned with the reasoning-baseline contrast than directness and curvature variability.

\section{Robustness Checks}
\label{app:robustness}

\subsection{Boundary policy and segmentation}
\label{app:boundary_segmentation}

Because reasoning models often expose explicit delimiters while baselines do not, we evaluate whether boundary policy could mechanically induce the observed coupling pattern.

\begin{table*}[t]
\centering
\caption{Boundary-detection rates by model tier and domain. Values are fractions of traces using each boundary source.}
\label{tab:boundary_detection_rates}
\small
\begin{tabular}{ll}
\toprule
Setting & Rates \\
\midrule
Pipeline tier, Codeforces & \texttt{</think>} $0.369$; code fence $0.604$; fallback $0.034$ \\
Pipeline tier, MATH       & \texttt{</think>} $0.391$; \texttt{\textbackslash boxed\{\}} $0.588$; fallback $0.021$ \\
Pipeline tier, SAT        & \texttt{</think>} $0.524$; \texttt{(UN)?SATISFIABLE} $0.456$; fallback $0.020$ \\
All tiers (fallback)      & API $0.035$; local $0.101$; pipeline $0.026$ \\
API-code detail           & No \texttt{<think>} tags; Claude Haiku 4.5 XML-tag rate $0.483$ on Codeforces \\
\bottomrule
\end{tabular}
\end{table*}

\begin{table*}[t]
\centering
\caption{Sensitivity of $\rho$ to boundary policy. Representative boundary-cut values at median $\tau$.}
\label{tab:boundary_sensitivity}
\small
\begin{tabular}{lccc}
\toprule
Model (Codeforces) & median $\tau$ & $\rho_{\mathrm{raw}}$ & $\rho_{\perp,\log T}$ \\
\midrule
DeepSeek-R1-7B & 0.93 & $-0.741$ & $+0.479$ \\
DeepSeek-R1-32B & 0.91 & $-0.755$ & $+0.404$ \\
QwQ-32B & 0.97 & $-0.704$ & $+0.411$ \\
Qwen-32B-Instruct & 0.38 & $-0.252$ & $-0.142$ \\
\bottomrule
\end{tabular}
\vspace{4pt}

\begin{tabular}{lccc}
\toprule
Model (MATH) & median $\tau$ & $\rho_{\mathrm{raw}}$ & $\rho_{\perp,\log T}$ \\
\midrule
DeepSeek-R1-7B & 0.82 & $-0.572$ & $+0.176$ \\
DeepSeek-R1-14B & 0.91 & $-0.538$ & $-0.033$ \\
QwQ-32B & 0.86 & $-0.523$ & $+0.030$ \\
Qwen-7B-Instruct & 0.98 & $-0.520$ & $-0.263$ \\
\bottomrule
\end{tabular}
\end{table*}

\begin{figure*}[t]
\centering
\includegraphics[width=\textwidth]{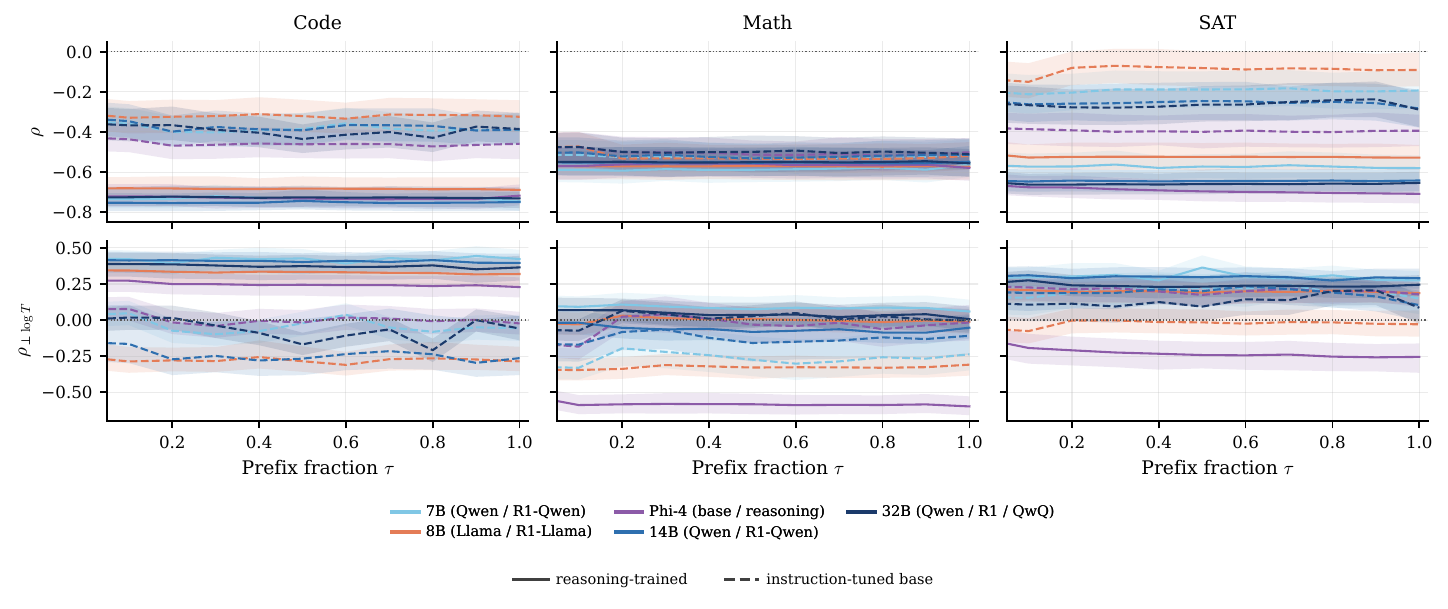}
\caption{Prefix and boundary sensitivity for directness-difficulty coupling. Prefix curves recompute corrected coupling using the first fraction of each generated solution segment; boundary variants compare full-output trajectories with answer-boundary cuts. Codeforces reasoning models show positive corrected coupling from the earliest measured prefixes, while MATH shows weaker and more gradual emergence. Boundary variants preserve the main Codeforces reasoning-baseline separation.}
\label{fig:directness_prefix_sensitivity}
\end{figure*}

Across tiers, API models do not emit \texttt{<think>} tags; API-code boundaries therefore rely on code fences, XML, or answer-style XML tags (Claude Haiku 4.5 XML-tag rate on Codeforces: $0.483$). Aggregate fallback rates are $0.035$ (API), $0.101$ (local), and $0.026$ (pipeline). Codeforces reasoning models consistently show raw-to-residual sign flips (DeepSeek-R1-32B: $-0.755 \to +0.404$; QwQ-32B: $-0.704 \to +0.411$); on math, residualized values are mostly small. Tagged-empty incidence is exactly zero across all 12 reasoning-model$\times$domain cases. Boundary-policy effects are also small for reasoning models in absolute terms ($\Delta_\mathrm{full}=|\rho_{\text{boundary}}-\rho_{\tau=1}|$: mean $0.009$, max $0.028$) and larger for baselines (mean $0.090$, max $0.221$).

Boundary-policy choice is therefore not the primary driver of the core code-domain result. Residualized coupling is substantially more stable than raw correlations across boundary variants. The only outlier is Phi-4-Reasoning on math under fixed-prefix variants ($\rho_{\perp,\log T}\approx -0.59$ to $-0.60$), which sits alongside the broader cross-model pattern rather than disturbing it.

\subsection{Layer sensitivity}
\label{app:layer_sensitivity}

\begin{table}[t]
\centering
\caption{Layer-stratified $\rho_\perp^D$ across the five sampled layers. Median, minimum, and maximum are taken over each model's five layer-specific values, then aggregated as a median over models within each domain $\times$ group cell. Counts use the pair-level convention.}
\label{tab:layer_robustness}
\small
\begin{tabular}{@{}llccc@{}}
\toprule
Domain & Group & Median (per-model layer median) & Min (across all layers) & Max (across all layers) \\
\midrule
Codeforces & Reasoning & $+0.41$ & $+0.26$ & $+0.48$ \\
Codeforces & Baseline  & $-0.05$ & $-0.14$ & $+0.08$ \\
MATH       & Reasoning & $+0.05$ & $-0.12$ & $+0.18$ \\
MATH       & Baseline  & $-0.05$ & $-0.35$ & $+0.07$ \\
SAT        & Reasoning & $+0.27$ & $-0.30$ & $+0.49$ \\
SAT        & Baseline  & $+0.23$ & $+0.10$ & $+0.33$ \\
\bottomrule
\end{tabular}
\end{table}

The qualitative Codeforces reasoning-baseline separation is present at every sampled depth: reasoning models show positive $\rho_\perp^D$ across all five sampled layers (median $+0.41$), while baselines remain near zero (median $-0.05$). On MATH, the effect is weak across layers in both groups. On SAT, the layer-stratified medians are similar between groups ($+0.27$ vs $+0.23$), reproducing the attenuated reasoning-specificity reported in the main text. Main figures report the median across layers; the qualitative conclusions do not depend on a single sampled layer.

\subsection{Null-label checks}
\label{app:null_label_checks}

As a null-label check, item difficulties are shuffled within each domain and model before recomputing corrected difficulty-geometry coupling. The Codeforces reasoning-model couplings lie in the tails of their null distributions, while the SAT pattern appears in both reasoning and baseline groups, matching the main analysis. This check shows that the observed rank associations are not typical of arbitrary difficulty assignments.

\subsection{Conditioning on correctness}
\label{app:correctness_conditioning}

\begin{table}[t]
\centering
\caption{Length-corrected directness-difficulty coupling within correctness strata. Group medians of $\rho_\perp^D$ over pair-level rows. Strata-size ranges are item counts retained per model after correctness filtering.}
\label{tab:correctness_stratified}
\small
\begin{tabular}{@{}llccc@{}}
\toprule
Domain & Group & Correct-only $\rho_\perp^D$ & Incorrect-only $\rho_\perp^D$ & Strata sizes (correct, incorrect) \\
\midrule
Codeforces & Reasoning & $+0.40$ & $+0.10$ & 257--362, 309--463 \\
Codeforces & Baseline  & $-0.11$ & $-0.00$ & 52--183, 459--498 \\
MATH       & Reasoning & $+0.03$ & $-0.02$ & 359--428, 141--236 \\
MATH       & Baseline  & $+0.05$ & $-0.03$ & 361--399, 162--333 \\
SAT        & Reasoning & $+0.51$ & $+0.31$ & 458--494, 240--389 \\
SAT        & Baseline  & $+0.23$ & $+0.16$ & 424--467, 346--481 \\
\bottomrule
\end{tabular}
\end{table}

Conditioning on correctness does not remove the main code-domain pattern. On Codeforces, reasoning models retain positive corrected directness coupling among correct traces and a smaller positive coupling among incorrect traces, while baselines remain near zero in both subsets. SAT shows the same attenuated pattern as the full analysis: reasoning models remain above baselines within each correctness subset, but baselines also show positive corrected coupling. MATH remains small in both groups. Because correctness still covaries with difficulty within these subsets, this check complements rather than replaces the full-domain estimates.

\subsection{Truncation and run-count stability}
\label{app:run_count_stability}

Truncation rates are low across pipeline tiers (Appendix~\ref{app:boundary_segmentation}). Removing truncated runs from the trajectory analyses preserves the qualitative reasoning-baseline separation reported in the main text.

R1-Distill-Qwen-7B on code uses 30 runs per problem. We compare trajectory statistics computed from 5 randomly sampled runs versus all 30 runs across 100 bootstrap resamples. The 30-run $\rho_\perp = +0.51$; the 5-run subsample mean is $+0.44$ (95\% CI: $[+0.37, +0.49]$). The intraclass correlation coefficient for problem-level mean directness is $\mathrm{ICC}(1,1) = 0.80$, indicating that approximately 80\% of directness variance is between-problem and that 5 runs per problem yield reasonably stable estimates.

\section{Probes and Interventions}
\label{app:probes_interventions}

\subsection{Linear difficulty decodability}
\label{app:linear_decodability}

We probe hidden states for linear difficulty decodability at two stages: at the final prompt token, before generation begins, and across a layer-by-position grid that spans the generated solution segment. These probes test whether the corrected geometry gap is mirrored by stronger linear access to difficulty; they do not test for nonlinear representations of difficulty, nor for differences in how equally accessible information is used during generation.

Prompt-stage probing extracts the hidden state at the last prompt token at every transformer layer for each of the eleven matched-pair models; the extraction reuses the forward pass that begins generation and keeps only the prompt-token states. Generation-stage probing samples each trace at ten evenly spaced positions (always including the first and last) and at every sampled layer. Hidden states are averaged across runs so that the effective sample size equals the number of unique problems, and trace length is residualized out of both the feature matrix and the difficulty target via OLS (Section~\ref{sec:confound}) before probing.

At each layer for the prompt stage and each (layer, position) cell for the generation stage, we standardize features and fit a Ridge probe~\citep{alain2017understanding}. RidgeCV selects $\lambda \in \{10^{-2}, 10^{-1}, 1, 10, 10^2, 10^3, 10^4\}$ by leave-one-out cross-validation, and generalization is estimated by 5-fold cross-validated $R^2$. A surface-feature floor uses the same Ridge probe on five descriptors of the input prompt (character length, word count, unique-token ratio, numeric literal count, sentence count) and gives $R^2 \approx 0.04$ on code and $0.08$ on math~\citep{hewitt2019control}. A permutation null shuffles difficulty labels 100 times per heatmap cell and reuses the precomputed decomposition of $\mathbf{X}_\mathrm{train}$ so that the procedure scales without refitting.

Probe scores are interpreted only within matched pairs, not as cross-family absolute quantities, because architectures differ in hidden dimension, tokenization, prompt format, and layer count. Within that scope, peak generation-stage $R^2$ on Codeforces runs from $0.22$ (R1-Distill-Llama-8B) to $0.37$ (Phi-4-Reasoning) for reasoning models, well above the surface floor. On SAT, the eleven matched-pair models span $0.16$ (Phi-4-Reasoning) to $0.50$ (Qwen2.5-14B-Instruct); reasoning models cluster in $0.16$--$0.33$ and matched baselines in $0.36$--$0.50$, reproducing the panel (c) gap of Figure~\ref{fig:prompt_generation_dissociation} at the per-model level.

Figure~\ref{fig:prompt_generation_dissociation} reports three reasoning-minus-baseline gaps over 18 pair-domain records (6 matched pairs $\times$ 3 domains): $\Delta R^2_{\mathrm{prompt}}$ from peak prompt-stage $R^2$ across layers, $\Delta R^2_{\mathrm{gen}}$ from peak generation-stage $R^2$ across the layer $\times$ position grid, and $\Delta\rho_\perp^D$ from the length-corrected directness-difficulty coupling. All three are signed within-pair differences.

\subsection{Temporal emergence of the geometric signal}
\label{app:temporal_emergence}

We analyze when the difficulty-geometry coupling emerges during generation by computing $\rho_\perp$ on progressively longer prefixes of each trace. All metrics are computed on the generated solution segment, consistent with the main trajectory analysis pipeline. The analysis covers all six reasoning-baseline pairs on Codeforces and MATH. In code, all six reasoning models show flat $\rho_\perp$ curves from the first 10\% prefix onward (all $|\Delta| < 0.05$ between 10\% and 100\%). In math, three of six models show building patterns: for R1-7B and R1-32B, the coupling builds from near zero to the full-trace value over the course of generation; Phi-4-Reasoning shows the most gradual trajectory, beginning negative and crossing zero after roughly two-thirds of generation.

\paragraph{Prefix directness.}
For each trace, hidden states are truncated to the generated solution segment (using the boundary detected as in Appendix~\ref{app:trajectory_extraction}) before computing prefix directness. At each prefix fraction $f \in \{0.1, 0.2, \ldots, 1.0\}$, we take the first $\lfloor f \cdot T_\mathrm{reasoning} \rfloor$ states (minimum 3 tokens), compute directness, average across runs per problem, and compute $\rho_\perp$ with bootstrap CIs ($n_\mathrm{boot} = 1{,}000$).

\begin{figure}[t]
\centering
\includegraphics[width=\textwidth]{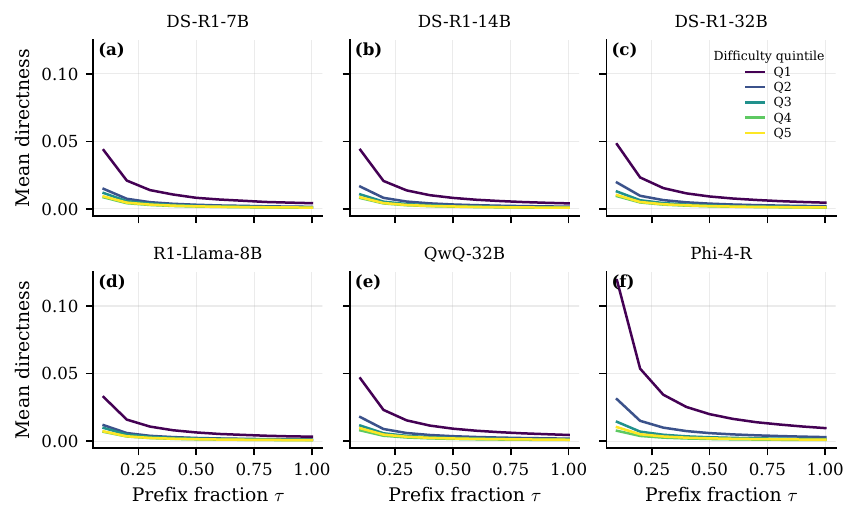}
\caption{Within-trace directness as a function of reasoning-prefix fraction for the six matched-pair reasoning models on Codeforces. Each panel plots mean directness across runs at ten prefix fractions $f \in \{0.1, 0.2, \ldots, 1.0\}$, stratified by pooled-IRT difficulty quintile (Q1 easiest in light yellow, Q5 hardest in dark purple). Codeforces curves are approximately flat from the 10\% prefix onward across all difficulty quintiles, consistent with the claim that the $\rho_\perp^D$ signal is established early in generation.}
\label{fig:temporal_dynamics}
\end{figure}

\paragraph{Within-trace segmentation.}
We divide each trace into non-overlapping windows of 100 tokens and compute per-window directness alongside behavioral density (strategy-shifting events per window, from sentence-level LLM-judge annotations). Wilcoxon signed-rank tests comparing shift-dense versus shift-sparse windows yield significant effects ($p < 0.05$) for 5 of 7 models tested in the code domain, with small effect sizes. The two null results (R1-Distill-Qwen-32B, $p = 0.35$; Llama-3.1-8B-Instruct, $p = 0.13$) suggest the within-trace signal is not universal. The difficulty-geometry coupling operates primarily at the whole-trace level, with modest within-trace contributions.

\subsection{Difficulty-direction interventions}
\label{app:difficulty_direction_interventions}

We probe four complementary tests of whether the linearly decodable difficulty direction carries the corrected geometric coupling: nullspace projection, iterative null-space erasure (INLP), variance-decomposition mediation, and matched-magnitude steering with random and orthogonal controls. All four tests run on DeepSeek-R1-Distill-Qwen-7B in code and mathematics, with Qwen2.5-7B-Instruct on code as a negative control.

\paragraph{Direction extraction.}
From the probing heatmap, we identify the (layer, position) cell with the highest $R^2$ and refit Ridge on the averaged, length-residualized data at that cell. We extract the weight vector $\mathbf{w}$ and transform it to the original feature space:
\begin{equation}
\label{eq:direction}
\hat{\mathbf{d}} = \frac{\mathbf{w} \oslash \mathbf{s}}{\|\mathbf{w} \oslash \mathbf{s}\|_2},
\end{equation}
where $\mathbf{s}$ is the vector of per-feature standard deviations and $\oslash$ denotes elementwise division. We project the training data onto $\hat{\mathbf{d}}$ and compute $\sigma_\mathrm{proj} = \mathrm{std}(\mathbf{X}\hat{\mathbf{d}})$; $\alpha = 1$ corresponds to a one-standard-deviation shift along the difficulty axis.

\paragraph{Nullspace projection.}
For each tested layer we project hidden-state trajectories into the nullspace of the probe direction $\hat{\mathbf{d}}$ and recompute $\rho_\perp^D$. Under the hypothesis that the probe direction carries the geometric signal, the projection should reduce $\rho_\perp^D$ relative to 100 random-direction controls of equal norm. Table~\ref{tab:nullspace_causal} reports the test on Codeforces and MATH at five layers, with Qwen2.5-7B-Instruct on Codeforces as a negative control. The drop is below $0.03\%$ at every tested layer for every model-domain, and the empirical $p$-value against the random-direction null exceeds 0.25 at every tested cell except Qwen2.5-7B-Instruct at layer 20. The probe direction does not carry the variance that produces $\rho_\perp^D$.

\begin{table}[h]
\centering
\caption{Nullspace projection test. $\rho_\perp$ is computed before and after projecting hidden-state trajectories into the nullspace of the probe direction. $p$ is the empirical $p$-value against 100 matched-norm random directions.}
\label{tab:nullspace_causal}
\small
\begin{tabular}{@{}llrrrrr@{}}
\toprule
Model & Domain & Layer & $\rho_\perp$ original & $\rho_\perp$ projected & $\Delta$ (\%) & $p$ \\
\midrule
R1-Distill-Qwen-7B & Codeforces & 0 & $+0.4517$ & $+0.4517$ & $-0.00$ & $0.257$ \\
 &  & 6 & $+0.4650$ & $+0.4650$ & $+0.00$ & $0.475$ \\
 &  & 13 & $+0.4829$ & $+0.4829$ & $+0.00$ & $0.337$ \\
 &  & 20 & $+0.4623$ & $+0.4623$ & $+0.00$ & $0.297$ \\
 &  & 27 & $+0.4666$ & $+0.4666$ & $+0.00$ & $0.347$ \\
\midrule
R1-Distill-Qwen-7B & MATH & 0 & $+0.1585$ & $+0.1585$ & $-0.00$ & $0.634$ \\
 &  & 6 & $+0.1756$ & $+0.1756$ & $-0.02$ & $0.960$ \\
 &  & 13 & $+0.1734$ & $+0.1734$ & $+0.00$ & $0.901$ \\
 &  & 20 & $+0.1677$ & $+0.1677$ & $+0.01$ & $0.446$ \\
 &  & 27 & $+0.1778$ & $+0.1778$ & $-0.00$ & $0.752$ \\
\midrule
Qwen2.5-7B-Instruct & Codeforces & 0 & $-0.0529$ & $-0.0530$ & $-0.26$ & $0.327$ \\
 &  & 6 & $-0.0113$ & $-0.0111$ & $+1.45$ & $0.901$ \\
 &  & 13 & $-0.0584$ & $-0.0583$ & $+0.23$ & $0.762$ \\
 &  & 20 & $-0.0618$ & $-0.0623$ & $-0.78$ & $0.010$ \\
 &  & 27 & $-0.0455$ & $-0.0455$ & $-0.00$ & $0.267$ \\
\bottomrule
\end{tabular}
\end{table}

\paragraph{INLP erasure.}
Iterative nullspace projection~\citep{ravfogel2020inlp} removes the top linear directions predictive of difficulty until cross-validated $R^2$ on difficulty drops below $0.02$. On Codeforces this takes three iterations and reduces probe $R^2$ from $0.215$ to $0.012$; on MATH it takes six iterations and reduces $R^2$ from $0.295$ to $0.017$. In both domains $\rho_\perp^D$ is unchanged to four decimal places after erasure (Codeforces: $0.4829 \to 0.4828$; MATH: $0.1734 \to 0.1737$). Removing all linearly decodable difficulty information leaves the corrected difficulty-geometry coupling intact.

\paragraph{Variance-decomposition mediation.}
We test whether the prompt-stage encoded representation of difficulty mediates the difficulty-geometry link via a variance-decomposition ratio: the variance of $\rho_\perp^D$ explained by the encoded difficulty (out-of-fold Ridge prediction) divided by the variance explained by the true difficulty label. On Codeforces the VD ratio is $0.171$ (bootstrap 95\% CI $[0.001, 3.159]$); on MATH the estimate is unstable ($5.048$, CI $[0.750, 371.387]$). The implied proportion mediated is negative in both domains ($-38\%$ on code, $-137\%$ on MATH), indicating that the prompt-stage encoding does not mediate the difficulty-geometry link.

\begin{figure}[t]
\centering
\includegraphics[width=\textwidth]{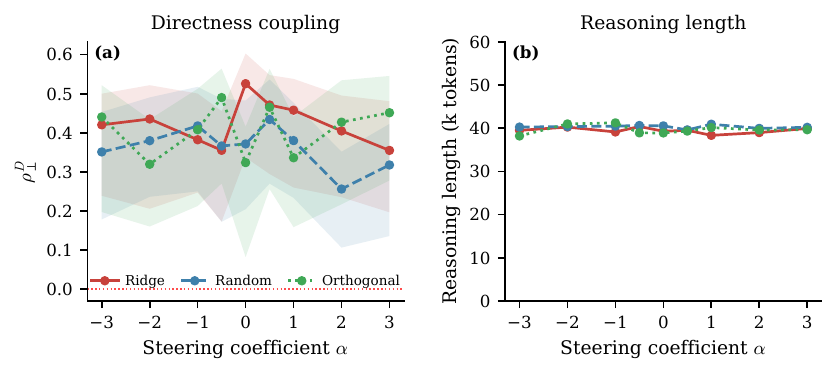}
\caption{Think-only steering on R1-Distill-Qwen-7B in Codeforces. Left: length-corrected coupling $\rho_\perp^D$ as a function of the steering coefficient $\alpha$ for three direction types (ridge probe, matched-norm random, orthogonal). Shaded bands are bootstrap 95\% CIs from 1{,}000 resamples of problems. Right: mean reasoning length per $\alpha$ for the same three directions. Ridge is statistically indistinguishable from the two controls in both panels, and reasoning length is invariant within $\pm 5\%$.}
\label{fig:causal_alpha_sweep}
\end{figure}

\paragraph{Think-only steering with controls.}
At each generation step within the \texttt{<think>} block, a forward hook modifies the hidden state at the target layer~\citep{turner2024steering, li2024inference}:
\begin{equation}
\label{eq:steering}
\mathbf{h}_t^{(\ell)} \leftarrow \mathbf{h}_t^{(\ell)} + \alpha \cdot \sigma_\mathrm{proj} \cdot \hat{\mathbf{d}}.
\end{equation}
We compare three direction types at nine $\alpha \in \{-3, -2, -1, -0.5, 0, +0.5, +1, +2, +3\}$: the ridge probe direction $\hat{\mathbf{d}}$, a random unit direction matched in norm, and a direction orthogonal to $\hat{\mathbf{d}}$. Figure~\ref{fig:causal_alpha_sweep} reports $\rho_\perp^D$ and reasoning length as functions of $\alpha$. The three direction types are statistically indistinguishable: ridge $\rho_\perp^D$ ranges from $0.36$ to $0.53$, random from $0.26$ to $0.43$, orthogonal from $0.32$ to $0.49$, with no monotonic dose-response in any of them. Reasoning length is invariant within $\pm 5\%$ across all $\alpha$ and all three direction types ($38{,}000$--$41{,}000$ tokens). The full nine-point grid is in Table~\ref{tab:steering_alpha_grid}.

\begin{table}[h]
\centering
\caption{Think-only steering: $\rho_\perp^D$ and mean reasoning length (tokens) at nine values of the steering coefficient $\alpha$, for ridge, random, and orthogonal directions.}
\label{tab:steering_alpha_grid}
\small
\begin{tabular}{@{}rrrrrrr@{}}
\toprule
 & \multicolumn{3}{c}{$\rho_\perp^D$} & \multicolumn{3}{c}{Reasoning length (tokens)} \\
\cmidrule(lr){2-4}\cmidrule(lr){5-7}
$\alpha$ & Ridge & Random & Orthogonal & Ridge & Random & Orthogonal \\
\midrule
$-3.0$ & $+0.420$ & $+0.351$ & $+0.441$ & $39{,}490$ & $40{,}248$ & $38{,}197$ \\
$-2.0$ & $+0.435$ & $+0.380$ & $+0.320$ & $40{,}245$ & $40{,}376$ & $40{,}976$ \\
$-1.0$ & $+0.382$ & $+0.418$ & $+0.408$ & $39{,}118$ & $40{,}468$ & $41{,}226$ \\
$-0.5$ & $+0.355$ & $+0.366$ & $+0.490$ & $40{,}386$ & $40{,}622$ & $38{,}913$ \\
$+0.0$ & $+0.526$ & $+0.372$ & $+0.324$ & $39{,}279$ & $40{,}531$ & $38{,}894$ \\
$+0.5$ & $+0.471$ & $+0.434$ & $+0.465$ & $39{,}506$ & $39{,}596$ & $39{,}320$ \\
$+1.0$ & $+0.458$ & $+0.380$ & $+0.336$ & $38{,}338$ & $40{,}906$ & $40{,}140$ \\
$+2.0$ & $+0.405$ & $+0.256$ & $+0.427$ & $38{,}969$ & $39{,}952$ & $39{,}564$ \\
$+3.0$ & $+0.355$ & $+0.318$ & $+0.451$ & $39{,}942$ & $40{,}188$ & $39{,}632$ \\
\bottomrule
\end{tabular}
\end{table}

\paragraph{Interpretation.}
The four tests are jointly inconsistent with a causal role for the linearly decodable difficulty direction. Difficulty is decodable from the representation but does not participate in the dynamics that produce difficulty-dependent geometric adaptation. In the vocabulary of \citet{schuessler2024aligned}, the probe direction is oblique to the computation rather than aligned with it. The geometric adaptation and the difficulty encoding appear to be independent consequences of reasoning training, observed in the same hidden states but not in the same subspace.

\subsection{Cross-model behavior-direction transfer}
\label{app:cross_model_steering}

The behavioral mediation analysis in Section~\ref{sec:behavior} identifies behavior categories that co-vary with the corrected geometric coupling within each model. We test whether the directions that encode those behaviors in one model transfer as causal handles in another. For each of the six judge categories (strategy shifting, uncertainty monitoring, self-correction, verification, problem restatement, subgoal decomposition) we extract a behavior direction from R1-Distill-Qwen-7B at layer 13 via length-residualized contrastive Ridge, then project the direction into the activation space of Qwen2.5-7B-Instruct via paired-token nearest-neighbor mapping on a held-out calibration set.

Diagnostic Spearman correlations between each transferred direction and the corresponding behavior rate in Qwen2.5-7B-Instruct are reported in Table~\ref{tab:cross_model_diag}. Four of the six directions are statistically indistinguishable from zero ($|\rho| < 0.04$, all $p > 0.1$); problem restatement is weakly significant ($\rho = -0.053$, $p = 8\times 10^{-3}$) but with negligible effect size; subgoal decomposition is the lone positive transfer ($\rho = 0.160$, $p < 10^{-15}$), and the same direction also correlates with item difficulty ($\rho = 0.132$). Active steering of Qwen2.5-7B-Instruct at nine $\alpha$ values along the four primary transferred directions produces no monotonic dose-response in $\rho_\perp^D$, accuracy, reasoning length, or per-behavior rates. Behavior directions extracted from one model do not transfer as functional controls in another, even between models of comparable scale.

\begin{table}[h]
\centering
\caption{Cross-model diagnostic. Spearman $\rho$ between a behavior direction extracted from R1-Distill-Qwen-7B at layer 13 and projected into Qwen2.5-7B-Instruct, against the matching behavior rate measured in Qwen2.5-7B-Instruct's own generations on Codeforces.}
\label{tab:cross_model_diag}
\small
\begin{tabular}{@{}lrr@{}}
\toprule
Transferred direction & $\rho$ vs target behavior rate & $p$ \\
\midrule
Strategy shifting & $+0.031$ & $0.119$ \\
Uncertainty monitoring & $-0.002$ & $0.931$ \\
Self-correction & $+0.004$ & $0.845$ \\
Verification & $-0.019$ & $0.353$ \\
Problem restatement & $-0.053$ & $0.008$ \\
Subgoal decomposition & $+0.160$ & $<10^{-15}$ \\
\bottomrule
\end{tabular}
\end{table}

\section{Behavioral Annotations}
\label{app:behavior}

\subsection{Annotation categories}

Three independent LLM judges classify each sentence of the generated solution segment. Behavioral categories are defined as follows:

\begin{itemize}
\item \textbf{Strategy shifting}: The model explicitly abandons or replaces its current approach (e.g., ``Let me try a different approach,'' ``Actually, this won't work because\ldots'').
\item \textbf{Uncertainty monitoring}: The model expresses doubt about its current reasoning, hedges a conclusion, or flags a potential error without yet changing strategy (e.g., ``I'm not sure this is right,'' ``Wait, let me check\ldots'').
\item \textbf{Self-correction}: The model identifies and corrects a specific error in its previous reasoning (e.g., ``No, that's wrong because\ldots,'' correcting a calculation).
\item \textbf{Verification}: The model checks a result by substitution, re-derivation, or testing (e.g., ``Let me verify by plugging in\ldots'').
\item \textbf{Problem restatement}: The model restates the problem, constraints, or goal without advancing the solution.
\item \textbf{Subgoal decomposition}: The model breaks the problem into named subproblems or explicitly sequences steps.
\end{itemize}

\subsection{Judge protocol and aggregation}
\label{app:judge_agreement}

Each judge receives the full reasoning trace with sentence boundaries pre-tokenized. The system prompt instructs the judge to classify each sentence into exactly one category. Majority-vote aggregation across the three judges produces the final label.

\paragraph{Inter-judge agreement.}
Mean pairwise Spearman $\rho$ on per-problem behavior rates: $\rho = 0.85$ (range: $0.81$--$0.89$ across category-judge pairs). Cohen's $\kappa$ on sentence-level labels: $0.72$ (substantial agreement). Disagreements concentrate on the self-correction / strategy-shifting boundary, where annotators differ on whether an error acknowledgment constitutes a correction or a strategy change.

\paragraph{Judges and metadata exposure.}
The three judges are Gemma-2-9B-IT, Llama-3.1-8B-Instruct, and Qwen2.5-7B-Instruct. Each judge labels the same sentence-segmented traces independently. Judges receive the category definitions of Appendix~\ref{app:behavior} as their system prompt and the sentence-segmented generated solution segment as their user prompt. Model identity, item difficulty, correctness outcome, trajectory metrics, and matched-pair labels are not included in the prompt; the domain is implicit because each judge run is per-domain.

\subsection{Residualized indirect-effect estimates}
\label{app:behavior_details}

This appendix reports residualized indirect-effect estimates that complement Section~\ref{sec:behavior}. All variables are residualized on $\log N$ and analyzed within model and domain.

\paragraph{R1-7B anchor result.}
For R1-Distill-Qwen-7B in the code domain (30 runs per problem, the richest data), strategy shifting and uncertainty monitoring have residualized indirect-effect proportions of $82\%$ and $98\%$, with bootstrap 95\% CIs excluding zero for both.

\paragraph{Cross-model results.}
Table~\ref{tab:mediation} reports indirect-effect proportions for all six reasoning models on Codeforces.

\begin{table}[h]
\centering
\caption{Residualized indirect-effect estimates: proportion of the difficulty-directness$_\perp$ co-variation that is statistically accounted for by each behavioral rate (code domain). All variables residualized on $\log N$. \textbf{Bold} = bootstrap CI excludes zero ($n_\mathrm{boot}=1{,}000$). $\dagger$~R1-7B uses 30 runs; others use 5. Values exceeding 100\% reflect partial suppression: when the direct effect is small relative to the indirect path and the two have opposite signs, the indirect-effect proportion can exceed unity.}
\label{tab:mediation}
\small
\begin{tabular}{@{}lcccc@{}}
\toprule
Model & Strat.\ Shift & Uncert.\ Mon. & Self-Corr. & Verific. \\
\midrule
R1-Distill-Qwen-7B$^\dagger$ & \textbf{82\%} & \textbf{98\%} & $-1\%$ & $-24\%$ \\
R1-Distill-Qwen-14B  & \textbf{71\%} & \textbf{91\%} & $<1\%$ & $-32\%$ \\
R1-Distill-Qwen-32B  & \textbf{145\%} & \textbf{141\%} & $-5\%$ & $-7\%$ \\
R1-Distill-Llama-8B  & \textbf{135\%} & \textbf{127\%} & $<1\%$ & $-9\%$ \\
\midrule
QwQ-32B              & $28\%$ & $12\%$ & $-7\%$ & $-27\%$ \\
Phi-4-Reasoning      & $30\%$ & $-6\%$ & $-3\%$ & \textbf{85\%} \\
\bottomrule
\end{tabular}
\end{table}

\begin{figure}[!t]
\centering
\includegraphics[width=\linewidth]{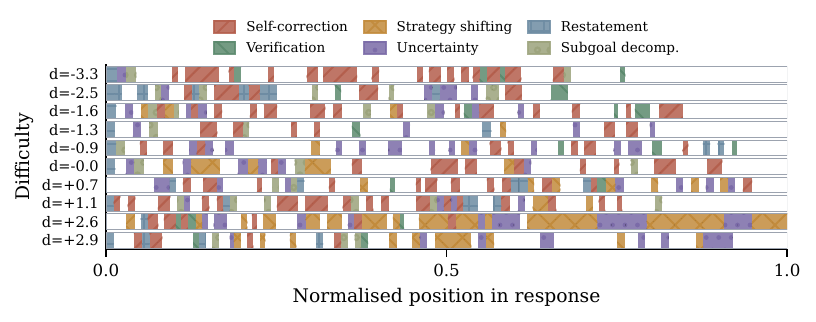}
\caption{\textbf{Where reasoning behaviors occur.} Each row is one
DeepSeek-R1-7B code trace; rows are ordered by pooled-IRT difficulty, with
two traces sampled from each difficulty quintile. The horizontal axis is
normalized character position in the full response (\texttt{0}: start of
$\langle$\texttt{think}$\rangle$; \texttt{1}: end of answer). Colored and
hatched spans show majority-vote sentence labels, requiring agreement from
at least two of three judges, for the six annotated behaviors. The gray tail
marks the post-$\langle$\texttt{/think}$\rangle$ answer segment. Strategy
shifting, self-correction, and uncertainty monitoring occur throughout the
trace, while verification is concentrated closer to the end; harder problems
show denser overlap of annotated behaviors.}

\label{fig:behavioral_stripe}
\vspace{-1em}
\end{figure}

The four R1-distilled models (sharing the DeepSeek-R1 teacher) show large indirect-effect proportions for both strategy shifting (4/4 with CIs excluding zero) and uncertainty monitoring (4/4). QwQ-32B and Phi-4-Reasoning, trained with different methods, show weaker indirect-effect estimates ($28\%$ and $30\%$ for strategy shifting; neither CI excludes zero with five runs). Phi-4-Reasoning instead shows a strong verification effect ($85\%$), a pattern absent in the distilled models. Estimates exceeding 100\% reflect partial suppression: when the direct effect ($c'$) is small relative to the indirect pathway ($ab$) and the two have opposite signs, the indirect-effect proportion exceeds unity.

Behavioral annotations and corrected geometry are derived from the same generated traces. These estimates therefore describe co-variation between observable reasoning behaviors and trajectory geometry; they do not establish that the annotated behaviors cause the geometric signal. The strongest co-variation is with strategy shifting and uncertainty monitoring in R1-distilled models, marking computational reorientation rather than continued elaboration along a fixed approach.

\paragraph{Within-trace segmentation.}
Within-trace segmented analysis (dividing traces into 100-token windows and testing whether shift-dense windows show different directness than shift-sparse windows) yields mixed results: 5 of 7 models tested show significant within-trace effects (Wilcoxon $p < 0.05$), but the effect sizes are small relative to the between-problem signal. The two null results (R1-Distill-Qwen-32B, $p = 0.35$; Llama-3.1-8B-Instruct, $p = 0.13$) suggest the within-trace signal is not universal. The indirect-effect estimate is dominated by between-problem variation. Figure~\ref{fig:behavioral_stripe} visualises where these behaviors fire across ten DeepSeek-R1-7B code traces sampled across the difficulty range.

\section{Data and Code Availability}
\label{app:data_code_availability}

The trajectory archive is available at \url{https://huggingface.co/datasets/gjoelbye/cot-hidden-state-trajectories}, with analysis code at \url{https://github.com/gjoelbye/reasoning-trajectory-geometry}. The archive pairs 1{,}500 calibrated items from Codeforces, MATH, and SATBench with generated traces for 32 models per domain, and includes sampled hidden-state activations for the eleven matched-pair models. Data are stored as problem parquets, trace parquets, and HDF5 activation archives, so subsets can be downloaded independently. The repository also includes notebooks that reproduce all tables and figures from pre-aggregated parquet files, without requiring the full $\sim$3 TB activation archive.



\end{document}